\documentclass[runningheads]{llncs}

\usepackage{eccv}

\usepackage{eccvabbrv}

\usepackage{graphicx}
\usepackage{booktabs}

\usepackage[accsupp]{axessibility}  

\usepackage[pagebackref,breaklinks,colorlinks,citecolor=eccvblue]{hyperref}

\usepackage{orcidlink}
\usepackage{adjustbox}

\usepackage{longtable}
\usepackage{bm}
\usepackage{wrapfig, blindtext}
\usepackage{multirow}
\usepackage{amssymb}
\usepackage{amsmath}

\begin{document}

\title{Look-Around Before You Leap: High-Frequency Injected Transformer for Image Restoration} 

\titlerunning{HIT}

\author{Shihao Zhou\inst{1,2} \and
Duosheng Chen\inst{1} \and
Jinshan Pan\inst{3} \and
Jufeng Yang\inst{1,2} }

\authorrunning{Zhou et al.}

\institute{VCIP \& TMCC \& DISSec, College of Computer Science, Nankai University \and
Nankai International Advanced Research Institute (SHENZHEN· FUTIAN) \and
School of Computer Science and Engineering, Nanjing University of Science and Technology}

\maketitle

\begin{abstract}
Transformer-based approaches have achieved superior performance in image restoration, since they can model long-term dependencies well. 
However, the limitation in capturing local information restricts their capacity to remove degradations. 
While existing approaches attempt to mitigate this issue by incorporating convolutional operations, the core component in Transformer, \ie, self-attention, which serves as a low-pass filter, could unintentionally dilute or even eliminate the acquired local patterns. 
In this paper, we propose \textbf{HIT}, a simple yet effective \textbf{H}igh-frequency \textbf{I}njected \textbf{T}ransformer for image restoration. 
Specifically, we design a window-wise injection module (WIM), which incorporates abundant high-frequency details into the feature map, to provide reliable references for restoring high-quality images. 
We also develop a bidirectional interaction module (BIM) to aggregate features at different scales using a mutually reinforced paradigm, resulting in spatially and contextually improved representations. 
In addition, we introduce a spatial enhancement unit (SEU) to preserve essential spatial relationships that may be lost due to the computations carried out across channel dimensions in the BIM. 
Extensive experiments on $\textbf{9}$ tasks (real noise, real rain streak, raindrop, motion blur, moir{\'e}, shadow, snow, haze, and low-light condition) demonstrate that HIT with linear computational complexity performs favorably against the state-of-the-art methods. 
{The source code and pre-trained models will be available at \href{https://github.com/joshyZhou/HIT}{https://github.com/joshyZhou/HIT}}.
\end{abstract}

\section{Introduction}
\label{sec:intro}

Image restoration aims to recover clear images by removing undesired degradation from input~\cite{lu2022needs}. 
Significant progress has been made due to the use of kinds of convolutional neural network (CNN) architectures~\cite{chen2022simple,pan2017deblurring,zhang2021pami}. 
However, the CNN-based methods are limited in modeling global contexts, which tends to negatively impact high-quality image restoration.

A newly proposed architecture, \ie, Transformer \cite{vaswani2017attention}, has attracted much attention from the vision and learning communities owing to its remarkable ability to capture long-range relations among distant pixels. 
Recent approaches \cite{wang2022uformer,chen2021pre,zamir2022restormer} successfully apply Transformer to image restoration, by reducing the quadratically grown complexity of vanilla Vision Transformers~(ViT)~\cite{iclr2021_vit}, \ie, $\mathcal{O}(N^2)$, where $N$ is the number of pixels (tokens). 
However, the limited capability of Transformer in aggregating local information~\cite{iccv2021conformer,d2021convit} remains a grand challenge. 
Note high-quality image restoration requires modeling both global and local information, and clear images typically contain global structures and rich local details. 
It is obviously insufficient to employ global dependencies alone for removing degradations, because the lack of high-frequency local information, which serves as guidelines to provide a reliable reference~(\eg, edges and textures), hinders these models from recovering fine details. 

\begin{figure*}[t]
\scriptsize
\centering
\begin{tabular}{ccc}
\begin{adjustbox}{valign=t}
\begin{tabular}{ccc}
\includegraphics[width=0.325\textwidth]{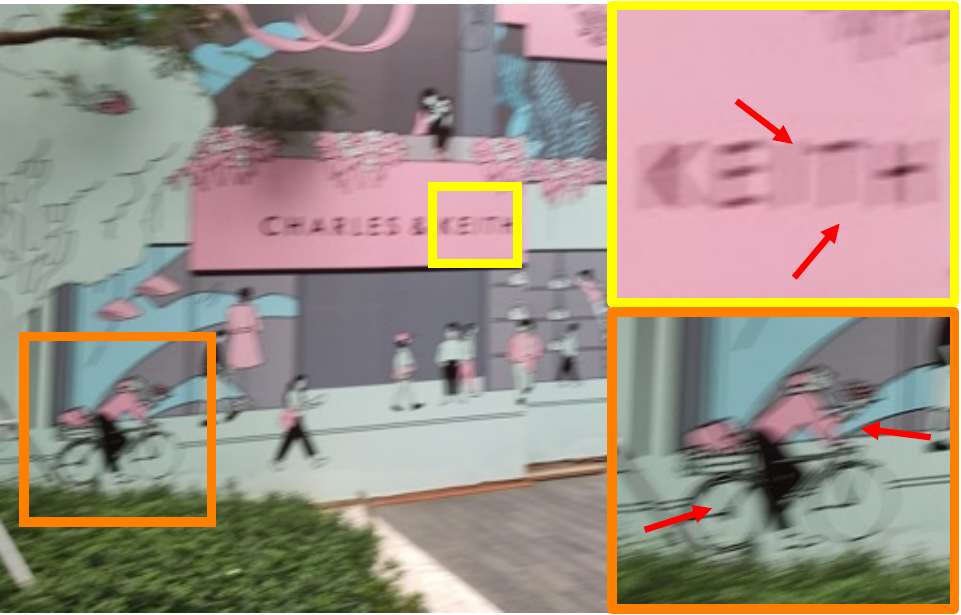} \hspace{-1.mm} &
\includegraphics[width=0.325\textwidth]{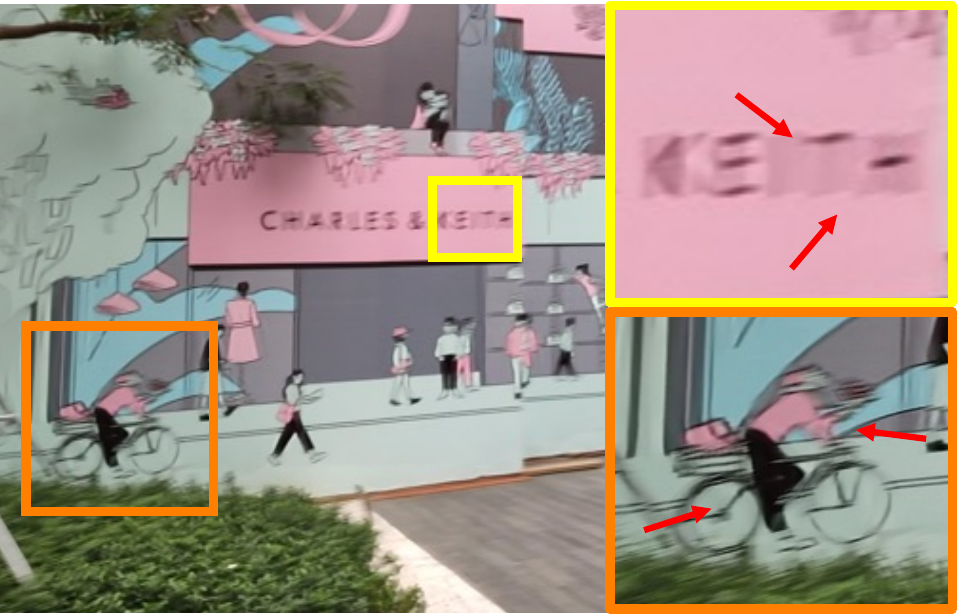} \hspace{-1mm} &
\includegraphics[width=0.325\textwidth]{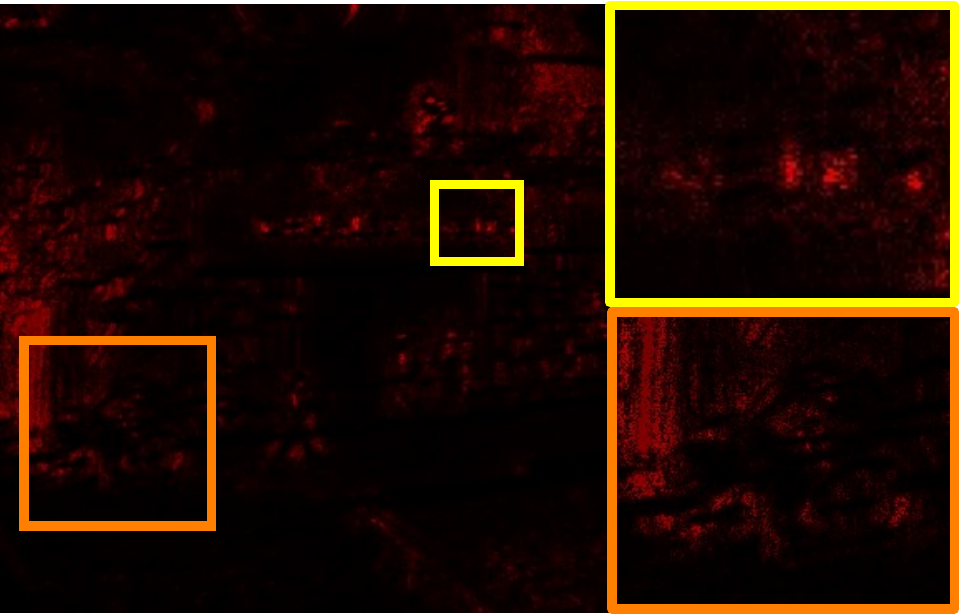} \hspace{-1.1mm} 
\\
\vspace{-0.35mm}
(a) Blurry Image \hspace{-1.5mm} &
(c) Uformer~\cite{wang2022uformer} \hspace{-2mm} &
(e) IG for Uformer~\cite{wang2022uformer}  \hspace{2.mm} 
\\
\includegraphics[width=0.325\textwidth]{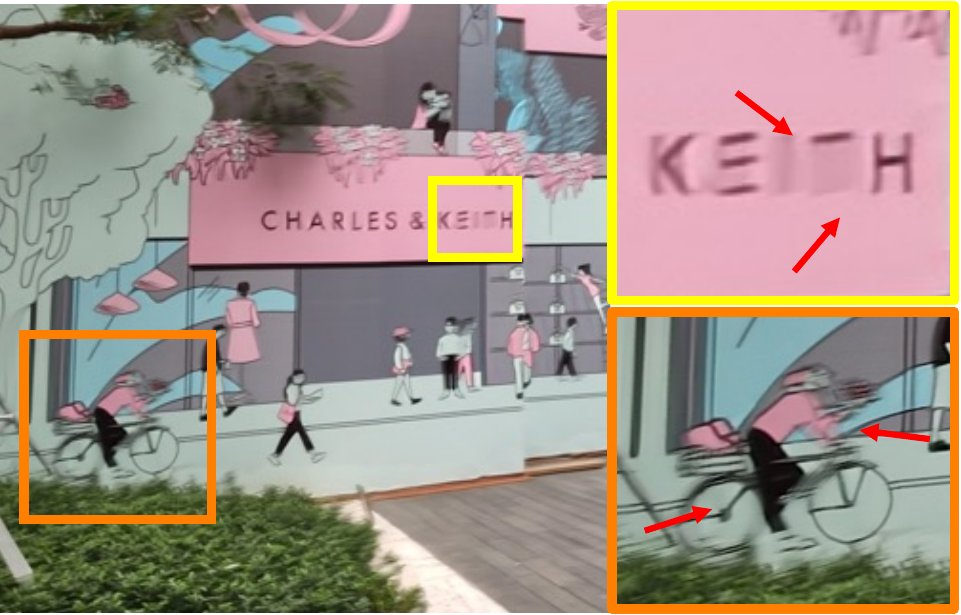} \hspace{-1.mm} &
\includegraphics[width=0.325\textwidth]{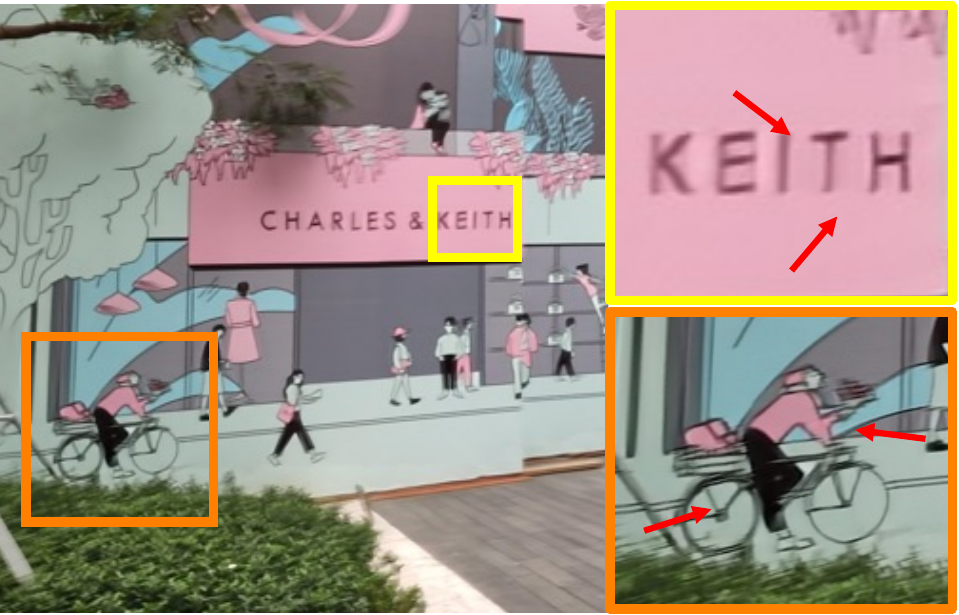} \hspace{-1mm} &
\includegraphics[width=0.325\textwidth]{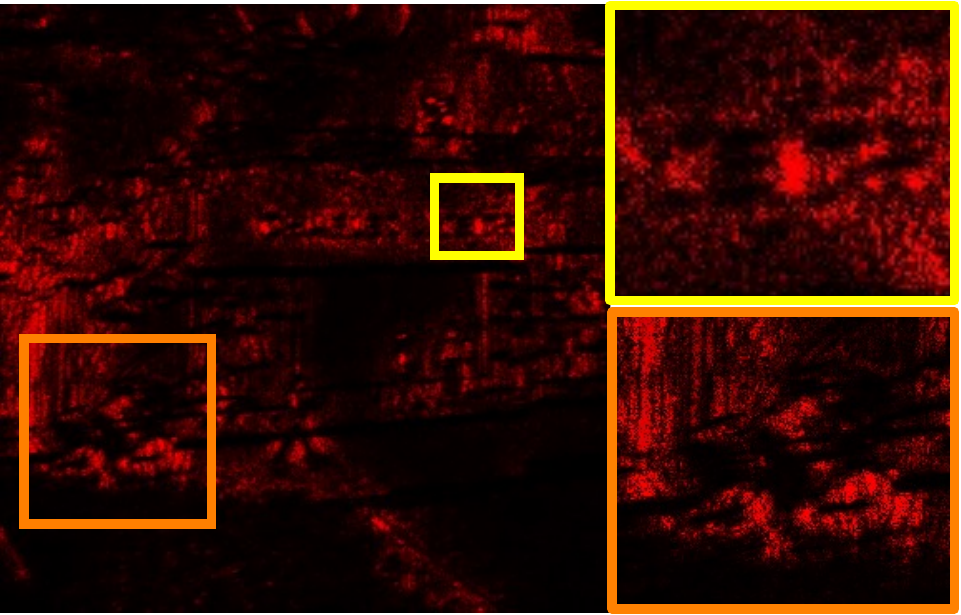} \hspace{-1.1mm}
\\ 
\vspace{-0.35mm}
(b) NAFNet~\cite{chen2022simple} \hspace{-2.5mm} &
(d) HIT (ours) \hspace{-3mm} &
(f) IG for HIT (ours) \hspace{-5mm}
\end{tabular}
\end{adjustbox}
\end{tabular}
\vspace{-2.5mm}
\caption{Image deblurring on the RWBI~\cite{zhang2020deblurring} dataset. 
Compared with the state-of-the-art approaches ((b) and (c)), the proposed HIT can generate clearer images as shown in (d).
Moreover, (e) and (f) denote the attribution maps of Uformer and HIT by using Integrated Gradients~(IG)~\cite{IG_17ICML}, where the pixel is activated when it contributes to the restoration result. 
}
\label{pic:rwbl}
\vspace{-3.5mm}
\end{figure*}

To mitigate this issue, a common solution is to introduce convolutional operations and learn convolution-like features~\cite{wang2022uformer,zamir2022restormer,eccv2022_Stripformer}. 
By incorporating depth-wise convolution into the Feed-Forward Network (FFN) of each Transformer block, local interactions within the feature map are enabled. 
The convolutional operation performs a weighted sum of neighborhoods, yielding feature representations focused on local correlations. 
When self-attention is applied to the convolved feature map, attention weights redistribute importance among pixels of the entire feature map, potentially reducing the emphasis on local patterns. 
In other words, the fine local patterns initially captured by the convolution layer may inadvertently be diluted or even eliminated, especially in deep layers of the network. 
As a result, it is hard to learn the desired fine high-frequency information within Transformer-based architecture due to its low-pass filter nature~\cite{park2022how}. 
In order to illustrate this, we utilize an attribution method, \ie, Integrated Gradients~\cite{IG_17ICML}, to discriminate which pixels contribute to the final prediction. 
As shown in Figure~\ref{pic:rwbl} (c), local details and structures were not sufficiently deblurred on the highly textured regions obtained by Uformer~\cite{wang2022uformer}, such as the numbers and the bicycle (zoom-in yellow and orange boxes).
When analyzing the corresponding attribution area in Figure~\ref{pic:rwbl} (e), we notice a lack of activations of neighboring pixels around the characters and the bike. 
This demonstrates that the diluted local cues can negatively impact the restoration of high-frequency details. 

In this paper, we propose a Transformer-based approach aiming at modeling local correlations for better image restoration.
The key idea of our \textbf{HIT} is \textbf{H}igh-frequency \textbf{I}njection in \textbf{T}ransformer with the proposed window-wise injection module (WIM). 
Different from existing methods that require a single model to capture both high-frequency and low-frequency information, we partition the learning targets into two categories: CNN handles low-level fine details, while Transformer tackles long-range dependencies. 
More specifically, we deploy a CNN-based extractor to generate high-frequency features thanks to its basic high-pass filter-like convolution operator~\cite{park2022how} and residual learning~\cite{pan2018learning}, and subsequently inject extracted features into Transformer in a window-wise fashion. 
In this way, our approach obtains plentiful high-frequency information while allowing the Transformer architecture to concentrate on modeling long-range relationships. 
Meantime, the global image contexts in the hierarchical features~\cite{zeiler2014visualizing} extracted by CNN, which plays a complementing role, provide a comprehensive picture of the image and alleviate the potential loss of global structure due to the window split strategy. 
As shown in Figure~\ref{pic:rwbl} (f), compared to existing methods, \eg,~\cite{wang2022uformer}, HIT effectively enhances high-frequency information without sacrificing the large receptive filed benefit of the Transformer. 
Moreover, towards preventing most useful high-frequency information from being diluted by the subsequent repeated self-attention mechanism, which serves as a low-pass filter~\cite{park2022how}, we tailor two schemes.
The first is to cut off the attention mechanism within the encoder part, which remains the FFN alone to deal with the information flow. 
The other one is to develop a bidirectional interaction module (BIM) to guide the feature integration for eliminating content information loss in the decoder part. 
The proposed BIM facilitates a two-way exchange of information between features at different scales, enabling each feature to benefit from the other's complementary characteristics. 
This bidirectional process involves calculating cross-attention from high-resolution features with fine details to semantically rich low-resolution representations and then in reverse, resulting in a spatially and contextually improved representation. 
Furthermore, we introduce a spatial enhancement unit~(SEU) to preserve the spatial information. 
The calculation in BIM, which is carried out across channel dimensions, could inadvertently lead to a loss of spatial context. 
In response, our SEU performs a convolution operation on the value projection in self-attention, and complements the aggregated feature of BIM with crucial spatial relations.

With the proposed modules, our HIT explores rich high-frequency information while restricting linear complexity. 
We perform comprehensive experiments on $\textbf{9}$ popular image restoration tasks, including image denoising, draining, deraindrop, deblurring, demoir{\'e}ing, deshadowing, desnowing, dehazing, and low-light image enhancement. 
Extensive experimental results show the effectiveness of our model. 
The main contributions of this work are threefold. 
(1) We propose an effective model, \ie,  HIT,  which leverages a CNN-based extractor to capture fine details, while ensuring that the Transformer focuses on modeling global context. 
This distinctive design enhances high-frequency information while maintaining the large receptive field benefit of the Transformer, thus facilitating high-quality image restoration. 
(2) We develop a window-wise injection module (WIM) to integrate high-frequency information into separate windows of the feature map. 
Towards keeping the most useful local cues can be met in deep layers, a bidirectional interaction module (BIM) is used to achieve spatially and semantically improved representations, in which a spatial enhancement unit~(SEU) is developed to ensure the crucial spatial details can be preserved in BIM. 
(3) We evaluate the proposed HIT on various tasks, showing that it achieves favorable performance.

\section{Related Work}
\label{sec:relatedWork}
\noindent \textbf{Image Restoration.}
Over the past decades, CNN-based methods~\cite{pami22_panXinggang,fan2020neural} have offered a preferable solution to the image restoration task, compared to the traditional approaches~\cite{cho2009fast,fergus2006removing}. 
By learning an optimal mapping function from low-quality images to high-quality ones, CNN-based architectures achieve impressive performance on various restoration tasks, including image denosing~\cite{pami22_sr_noise,pami23_wang_noise,chen2021pre}, deblurring~\cite{pami22_noise_blur_kerihuan,pan2017deblurring,Zhao_2023_CVPR}, deraining~\cite{pami23_rain_huang,DRSformer,10035447}, demoir{\'e}ing~\cite{rdnet_tmm22,zhang2023realtime,tpami22_zheng_demoire}, \etc. 
%
Since the introduction of CNN~\cite{sun2015learning,cvpr2021cho}, a surge of approaches have considered deeper and wider architecture designs~\cite{pami22_blur_renWenqi,pami21_blur_liujun} to explore global cues and further improve performance. 
Meanwhile, some works~\cite{tip23_songxibin,zamir2022restormer} introduce spatial and channel attention mechanisms to get better performance by forcing the model to focus more on relevant information.  
More architecture designs can be found in NTIRE challenge reports~\cite{Nah_2021_CVPR,Lugmayr_2022_CVPR} and recent surveys~\cite{zhang2022deep,pami22_survey_enhancement}. 
Besides, some works explore All-In-One image restoration~\cite{potlapalli2023promptir,TAPE-Net_eccv22}, which is out of the scope of this work. 

Recently, since Transformer~\cite{vaswani2017attention} has achieved great success in various natural language processing tasks, many works attempt to apply it in computer vision tasks~\cite{zheng2022cross,DRSformer,Guo_2023_ICCV}. 
Specifically, for image restoration, IPT~\cite{chen2021pre} first utilizes the vanilla Transformer as the backbone and obtains competitive results. 
Nevertheless, there remain concerns since it highly relies on large training data to fit a large number of parameters.
Stripformer~\cite{eccv2022_Stripformer} designs the novel intra- and inter-strip attention to form a token-efficient transformer. 
Restormer~\cite{zamir2022restormer} models global relations across channel dimensions to reduce complexity. 
However, Stripformer still needs high complexity (\ie, $\mathcal{O}(HW(H+W))$) while channel-wise attention in Restormer may lose necessary spatial information. 
On the other hand, some works~\cite{wang2022uformer,iccv2021_swinIR,li2023grl} leverage a window-based strategy~\cite{liu2021swin} to achieve linear complexity. 
Even though these works have made clear improvements, however, the insufficient local detail issue still limits the performance of the Transformer. 

\noindent\textbf{Frequency Components based Image Restoration.}
Apart from mining relations in the spatial domain, some works attempt to design networks solving various degradation removal tasks from a frequency perspective. 
To be specific, transformation tools, such as Fourier transform or wavelet, are employed by some works~\cite{guo2023spatial,zou2021sdwnet,Chen_2021_ICCV} to decompose features into different frequency bands. 
Unfortunately, few of them aim to use high-frequency information to improve transformers. 
How to effectively explore high-frequency information to ensure the Transformer-based methods can model both high-frequency local details and low-frequency non-local structures for better image restoration is not trivial, as Transformer-based mechanisms do not model high-frequency information well~\cite{park2022how}. 
Different from these methods, our HIT employs WIM using a $split$-$align$-and-$fuse$ strategy to emphasize the indispensable role of local details in the feature maps, which are then fused along the channel dimension. 
Besides, we cut off the self-attention mechanism in the encoder of the model to prevent the fine details from being diluted, and develop a BIM to deliver a spatially and contextually improved representation for the decoder part. 

\begin{figure*}[t]
\centering
\includegraphics[width=0.98\linewidth]{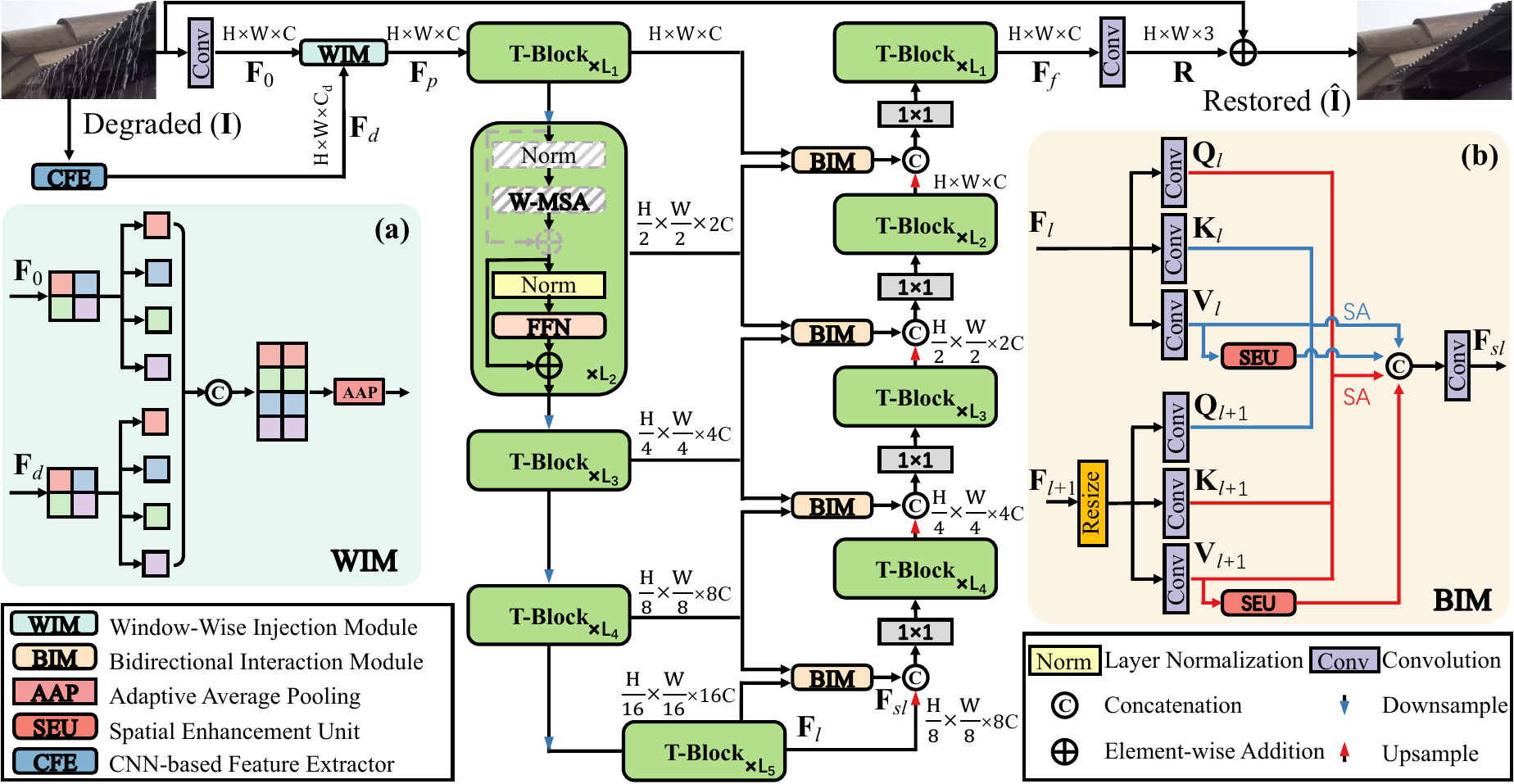}
\vspace{-2.5mm}
\caption{Overview of HIT. 
It consists of a U-shaped architecture and two modules: 
\textbf{(a)} Window-wise Injection Module~(WIM) that fuses local cues in separate windows of the feature map. 
\textbf{(b)} Bidirectional Interaction Module~(BIM) that aggregates features at different scales to achieve spatially and semantically improved representations. 
T-Block is short for Transformer Block.
W-MSA and FFN represent window-based multi-head self-attention~\cite{liu2021swin} and Feed-Forward Network\cite{li2021localvit}.
}
\label{pic:overall}
\vspace{-3.5mm}
\end{figure*}

\section{Proposed Method}
Figure \ref{pic:overall} shows an overview of our HIT model. It comprises a U-shaped architecture with two proposed modules: the window-wise injection module (\textbf{WIM}), which is described in detail in Section~\ref{method:first}, and the bidirectional interaction module~(\textbf{BIM}), which is presented in Section~\ref{method:second}.
%
\subsection{Overall Pipeline}
\label{subsec: overall}
Given a degraded image $\mathbf{I} \in \mathbb{R}^{H\times W\times 3}$, HIT first adopts a convolution layer to extract low-level feature $\mathbf{F}_{0} \in \mathbb{R}^{H\times W\times C}$, where $H,~W,~C$ denote the height, width and number of channels separately. 
The low-level feature is then taken into WIM, where the rich local feature $\mathbf{F}_{d} \in \mathbb{R}^{H\times W\times C_d}$ is encoded into the network (see Section \ref{method:first} for details). 
Next, the feature $\mathbf{F}_p \in \mathbb{R}^{H\times W\times C}$ processed by WIM is fed into $L$-level encoder-decoder parts and outputs the refined feature $\mathbf{F}_f \in \mathbb{R}^{H\times W\times C}$. 
Each level of encoder and decoder shares the same window-based multi-head self-attention block ($\textbf{W} \mbox{-} \textbf{MSA}$)~\cite{liu2021swin}. 
Following the attention block, a Feed-Forward Network~($\textbf{FFN}$), as pioneered works \cite{li2021localvit}, is employed. 
To be specific, in the encoder, the input feature is progressively processed by each Transformer block and generates the intermediate feature $\mathbf{F}_l \in \mathbb{R}^{\frac{H}{2^l}\times \frac{W}{2^l}\times 2^lC}$ at $l$-th depth, which is formulated as:
\begin{equation}
\begin{aligned}
        \hat{\mathbf{F}}_{l} &=\textbf{W} \mbox{-} \textbf{MSA}(\textbf{LN}( \mathbf{F}_{l-1}))+\mathbf{F}_{l-1},\\
        \mathbf{F}_{l} &= \textbf{FFN}(\textbf{LN}( \hat{\mathbf{F}}_{l}))+\hat{\mathbf{F}}_{l},
\label{eq:encoder}
\end{aligned}
\end{equation}
where $\textbf{LN}$ is layer normalization. 
Afterward, a convolution layer is used to resize the feature map. 

In the decoder, each level contains a Transformer block similar to the encoder part, except the convolution layer and the proposed BIM~(see Section~\ref{method:second}
for details). 
Specifically, the convolution layer in the decoder part performs feature up-sampling. 
Here, the input feature to the Transformer block at the $L$ stage in the decoder can be represented as:
\begin{equation}
\begin{aligned}
        \mathbf{F'}_{l} ={\rm Conv}_{1\times 1}([\mathbf{F}_{l}, \mathbf{F}_{sl}]) ,
\label{eq:decoder}
\end{aligned}
\end{equation}
where $[\cdot,\cdot]$ denotes the concatenation, Conv$_{1\times 1}(\cdot)$ is the 1$\times$1 convolution, $\mathbf{F}_{l} \in \mathbb{R}^{\frac{H}{2^l}\times \frac{W}{2^l}\times 2^lC}$ is the up-sampled feature, and $\mathbf{F}_{sl} \in \mathbb{R}^{\frac{H}{2^l}\times \frac{W}{2^l}\times 2^lC}$ is the output feature of BIM. 

After the $L$-level encoder-decoder architecture, we obtain the refined feature $\mathbf{F}_f \in \mathbb{R}^{H\times W\times C}$.
A convolution layer is then applied to generate a residual image $\mathbf{R} \in \mathbb{R}^{H\times W\times 3}$. 
Finally, the restored image is obtained by adding the degraded image: $\mathbf{\hat{I}} = \mathbf{I} + \mathbf{R}$. 
Similar to~\cite{wang2022uformer}, we adopt the commonly used Charbonnier loss~\cite{charbonnier1994two} to train the whole network: $\ell({\mathbf{I '}} ,\mathbf{\hat{I}})=\sqrt{\Vert {\mathbf{I '}} - \mathbf{\hat{I}}\Vert^{2} +\epsilon^{2}}$,
where ${\mathbf{{I'}}}$ denotes the ground-truth image and $\epsilon$ is set to $10^{-3}$.

\vspace{-3mm}
\subsection{Window-wise Injection Module}
\label{method:first}
%
%
Unlike existing methods that either adopt convolution~\cite{iccv2021conformer,chen2021transunet,Liu_2023_CVPR_TCM} or attention-based~\cite{cvpr2022mobile} fusion paradigm, as shown in Figure~\ref{pic:overall} (a), we develop window-wise injection module~(\textbf{WIM}), using a $split$-$align$-and-$fuse$ strategy, to encode high-frequency information into feature map along the channel dimension.
%

\textbf{Split:} First, a pre-trained CNN feature extractor (\eg, ResNet, where the final fully connected layer from the original design is ignored, remaining the final feature vector as input representation) is applied to the degraded image $\mathbf{I}$ to generate a feature representation with abundant local correlations, denoted as $\mathbf{F}_d \in \mathbb{R}^{H\times W\times C_{d}}$. 
%
Next, the input feature $\mathbf{F}_{0}$ and $\mathbf{F}_d$ are divided into non-overlapping windows with the same size of $M \times M$, which results in the separated version $\mathbf{F}'_{0} \in \mathbb{R}^{\frac{HW}{M^2}\times M^2\times C}$ and $\mathbf{F}'_{d} \in \mathbb{R}^{\frac{HW}{M^2}\times M^2\times C_d}$, and the $i$-th window features are denoted as $\mathbf{F}_{0\_i} \in \mathbb{R}^{(M\times M)\times C}$ and $\mathbf{F}_{d\_i} \in \mathbb{R}^{(M\times M)\times C_{d}}$ separately. 
\textbf{Align:} These window features are concatenated along the channel dimension with the same index $i$, which generates intermediate tensor $\mathbf{F}_{inter}\in\mathbb{R}^{\frac{HW}{M^2}\times M^2\times (C+C_d)}$.  
\textbf{Fuse:} The processed features are then sent into the adaptive average pooling layer~(AAP) for consolidation. 
Overall, the WIM process can be denoted as:
\begin{equation}
\begin{aligned}
\\[-9.5mm]
        \mathbf{F}_{p} ={\rm AAP}([\textbf{WP}(\mathbf{F}_{0}), \textbf{WP}(\mathbf{F}_{d})]) ,
\\[-9.5mm]
\label{eq:FPM}
\end{aligned}
\end{equation}
where $\textbf{WP}$ represents the window partition strategy~\cite{liu2021swin}. 
$\mathbf{F}_{p} \in \mathbb{R}^{{H}\times W\times C}$ is the reshaped output feature of WIM.

\vspace{-3.mm}
\subsection{Bidirectional Interaction Module }
\vspace{-1.mm}
\label{method:second}
%
To prevent the injected high-frequency information from being diluted by the low-pass filter-like self-attention mechanism, we tailor two specific schemes. 
First, we cancel the self-attention in the encoder to ensure these local details can flow forward. 
Second, we design a bidirectional interaction module ~(\textbf{BIM}) for useful local relations that can be met in the decoder, as illustrated in Figure~\ref{pic:overall}~(b).

Given the input feature maps $\mathbf{F}_l\in\mathbb{R}^{H'\times W'\times C'}$ and $\mathbf{F}_{l+1} \in \mathbb{R}^{\frac{H'}{2}\times \frac{W'}{2}\times 2C'}$, we first resize $\mathbf{F}_{l+1}$ and obtain $\mathbf{\tilde{F}}_{l+1}\in\mathbb{R}^{H'\times W'\times 2C'}$. 
%
Next, we generate linear projections $\mathbf{Q}_{l}$, $\mathbf{K}_{l}$, $\mathbf{V}_{l}\in \mathbb{R}^{H'\times W'\times C'}$ from $\mathbf{F}_{l}$, and $\mathbf{Q}_{l+1}$, $\mathbf{K}_{l+1}$, $\mathbf{V}_{l+1} \in \mathbb{R}^{H'\times W'\times 2C'}$ from $\mathbf{\tilde{F}}_{l+1}$. 
%
We estimate the cross-scale self-attention by: 
\begin{equation}
\begin{aligned}
\\[-8.5mm]
&{\rm Att}(\hat{\mathbf{Q}}_{l+1}, \hat{\mathbf{K}}_{l}, \hat{\mathbf{V}}_{l})= {\rm SoftMax}(\frac{\hat{\mathbf{Q}}_{l+1} \hat{\mathbf{K}}_{l}}{\alpha})\hat{\mathbf{V}}_{l},\\
&{\rm Att}(\hat{\mathbf{Q}}_{l}, \hat{\mathbf{K}}_{l+1}, \hat{\mathbf{V}}_{l+1})= {\rm SoftMax}(\frac{\hat{\mathbf{Q}}_{l} \hat{\mathbf{K}}_{l+1}}{\alpha})\hat{\mathbf{V}}_{l+1},
\\[-8.5mm]
\label{eq:cross-attention}
\end{aligned}
\end{equation}
where $\hat{\mathbf{Q}}_{l+1}$, $\hat{\mathbf{V}}_{l+1}\in \mathbb{R}^{2C'\times H'W'}$, $\hat{\mathbf{K}}_{l+1}\in \mathbb{R}^{H'W'\times 2C'}$, $\hat{\mathbf{Q}}_{l}$, $\hat{\mathbf{V}}_{l}\in \mathbb{R}^{C'\times H'W'}$, $\hat{\mathbf{K}}_{l}\in \mathbb{R}^{H'W'\times C'}$, SoftMax$(\cdot)$ represents the softmax activation and $\alpha$ is the learnable scaling factor. Here we use the transposed scaled-dot-product attention to reduce the computational cost according to \cite{zamir2022restormer}. 
%
We keep the calculation paradigm from \cite{vaswani2017attention}, \ie, the queries from one feature while the keys and the values come from another one, to encourage the rich interaction among every pixel in the feature maps.
We fuse the above estimated scaled-dot-product attentions so that the aggregated features can better explore cross-scale information. 
\begin{equation}
\begin{aligned}
\\[-10.5mm]
\mathbf{F}_{sl} ={\rm Conv}_{1\times 1}([\text{Att}(\hat{\mathbf{Q}}_{l}, \hat{\mathbf{K}}_{l+1}, \hat{\mathbf{V}}_{l+1}),\text{Att}(\hat{\mathbf{Q}}_{l+1}, \hat{\mathbf{K}}_{l}, \hat{\mathbf{V}}_{l})]).
\\[-10.5mm]        
\label{eq:SPM}
\end{aligned}
\end{equation}

\begin{wrapfigure}{r}{0.4\linewidth}
\centering
\vspace{-10.5mm}
\includegraphics[width=\linewidth]{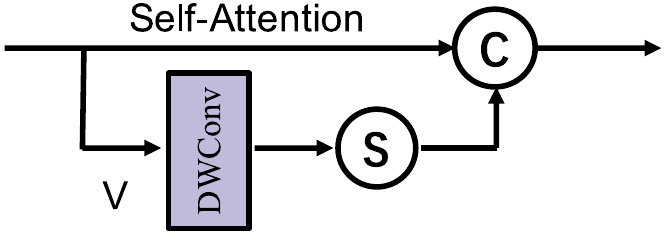}
\vspace{-7.mm}
\caption{Spatial Enhancement Unit. 
V stands for the Value projection in Self-Attention and DWConv is a depth-wise convolution. 
$\large{\copyright}$ denotes the concatenation operation and $\circledS$ is the softmax activation.} 
\label{pic:sem}
\vspace{-28mm}
\end{wrapfigure}


\noindent\textbf{Spatial Enhancement Unit.}
By calculating cross-covariance across channels, BIM results in the linear complexity of SA operation. 
Such channel-wise attention mechanisms \cite{zamir2022restormer}, however, may lose crucial spatial information. 
To address this issue, we design a unit to retain the indispensable spatial relations, namely spatial enhancement unit (SEU), as shown in Figure \ref{pic:sem}.

Take $\text{Attention}(\hat{\mathbf{Q}}_{l+1}, \hat{\mathbf{K}}_{l}, \hat{\mathbf{V}}_{l})$ as an example, the SEU is performed on the transposed results of the value $V$ by:
\begin{equation}
\begin{aligned}
        {\hat{\mathbf{V}}_{l}^{\prime}} =\sigma({\rm DWConv}_{3\times 3}(\hat{\mathbf{V}}_{l})),
\label{eq:SEM}
\end{aligned}
\end{equation}
where DWConv$_{3\times 3}(\cdot)$ is the 3$\times$3 depth-wise convolution, $\sigma$ is the activation function, and $\hat{\mathbf{V}}_{l} \in \mathbb{R}^{C'\times H'W'}$ is the transposed version of $V$, which is estimated from $\mathbf{F}_{l}$.
%
Then, we update Attention$(\cdot)$ with ${\hat{\mathbf{V}}_{l}^{\prime}}$ by:
{\begin{equation}
\begin{aligned}
{\rm Att}&(\hat{\mathbf{Q}}_{l+1}, \hat{\mathbf{K}}_{l}, \hat{\mathbf{V}}_{l})=[({\rm SoftMax}(\frac{\hat{\mathbf{Q}}_{l+1} \hat{\mathbf{K}}_{l}}{\alpha})\hat{\mathbf{V}}_{l},{\hat{\mathbf{V}}_{l}^{\prime}}].
\label{eq:reSA}
\end{aligned}
\end{equation}}
\textbf{Complexity Analysis.} Given the feature $\bm{F}_l\in \mathbb{R}^{{H}\times {W}\times C}$, the computational complexity of BIM is:
{\begin{equation}
\begin{aligned}
\mathcal{O}(\text{BIM})&=2(\mathcal{O}(\text{SEU})+\mathcal{O}(\text{SA}))=HWC\times(18+4C).
\label{eq:complexity}
\end{aligned}
\end{equation}}
%

\section{Experiments}
In this section, we begin with the experimental setup. 
Then we demonstrate the effectiveness of our method on five image restoration tasks~(\ie, denosing, deraining, low-light image enhancement, dehazing and deblurring). 
After that, we conduct ablation studies to verify the design contributions of each component. 
Due to the limited space, more numerical and visual results (\eg, deshadowing on ISTD~\cite{wang2018stacked}, desnowing on Snow100K~\cite{liu2018desnownet}, demoir{\'e}ing on TIP18~\cite{MSNet}, and deraindrop on AGAN-Data~\cite{qian2018attentive}) and detailed experimental settings are reported in supplementary materials.
\subsection{Experimental Setup}
\textbf{Metrics.}
%
%
%
%
%
%
%
We use the PSNR and SSIM metrics to evaluate the quality of each restored image when its ground truth image is available. 
Specifically, the metrics are applied in the RGB color space for most cases while calculated on the Y channel in YCbCr color space for deraining, following existing works~\cite{wang2022uformer,wang2020model}. 
For the reported results, the best and second best scores are \textbf{highlighted} and \underline{underlined}. 
%
%
For the evaluated methods, we report the results in their paper if provided~(\eg, DIL~\cite{DIL_2023_CVPR} for image denoising), otherwise we retrain the models with their publicly available code or evaluate with their pre-trained models~(\eg, NAFNet~\cite{chen2022simple}, trained on GoPro~\cite{cvpr2017_gopro}, evaluated on RealBlur~\cite{eccv2020real} for image deblurring). 
For others, we use the reported results in \cite{xiao2022image,wang2022uformer,10035447,MopNet}~(\eg, the result of Restormer~\cite{zamir2022restormer} on SPAD~\cite{wang2019spatial} for image deraining is from \cite{xiao2022image}).
%
\begin{figure*}[t]
\scriptsize
\centering
\begin{tabular}{ccc}
\hspace{-0.22cm}
\begin{adjustbox}{valign=t}
\begin{tabular}{c}
\includegraphics[width=0.185\textwidth]{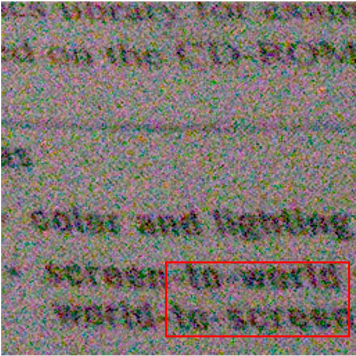}
\\
Noisy Image
\end{tabular}
\end{adjustbox}
\hspace{-0.10cm}
\begin{adjustbox}{valign=t}
\begin{tabular}{cccccc}
\vspace{0.4mm}
\includegraphics[width=0.196\textwidth]{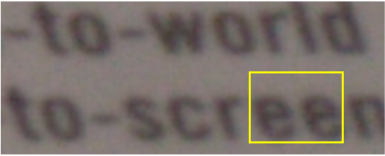} \hspace{-1mm} &
\includegraphics[width=0.196\textwidth]{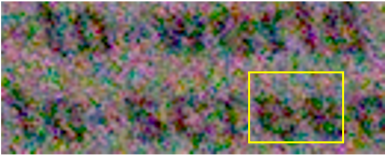} \hspace{-1mm} &
\includegraphics[width=0.196\textwidth]{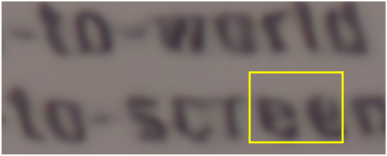} \hspace{-1mm} &
\includegraphics[width=0.196\textwidth]{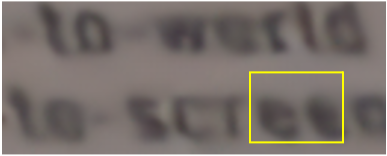} \hspace{-1mm} 
\\
\vspace{-0.8mm}
Reference \hspace{-1mm} &
Noisy \hspace{-1mm} &
CycleISP~\cite{zamir2020cycleisp} \hspace{-1mm} &
RIDNet~\cite{Anwar_2019_ICCV} \hspace{-1mm} 
\vspace{1.4mm}
\\
\includegraphics[width=0.196\textwidth]{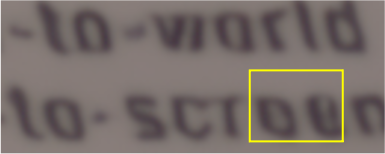} \hspace{-1mm} &
\includegraphics[width=0.196\textwidth]{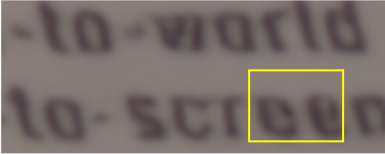} \hspace{-1mm} &
\includegraphics[width=0.196\textwidth]{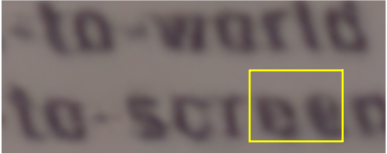} \hspace{-1mm} &
\includegraphics[width=0.196\textwidth]{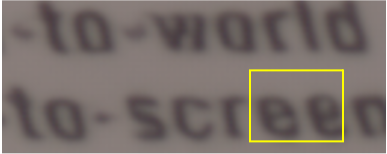} \hspace{-1mm} 
\\ 
MPRNet~\cite{zamir2021multi} \hspace{-1mm} &
MIRNet-v2~\cite{MIRNetv2} \hspace{-1mm} &
VDN~\cite{yue2019variational} \hspace{-1mm} &
\textbf{HIT-B} \hspace{-1mm}
\\
\end{tabular}
\end{adjustbox}
\end{tabular}
\vspace{-2mm}
\caption{Qualitative comparisons with SOTA methods on SIDD~\cite{ssid_2018} for denoising. }
\label{pic:denoise}
\vspace{-4mm}
\end{figure*}
\begin{figure*}[t]
\scriptsize
\centering
\begin{tabular}{ccc}
\hspace{-0.2cm}
\begin{adjustbox}{valign=t}
\begin{tabular}{c}
\includegraphics[width=0.185\textwidth]{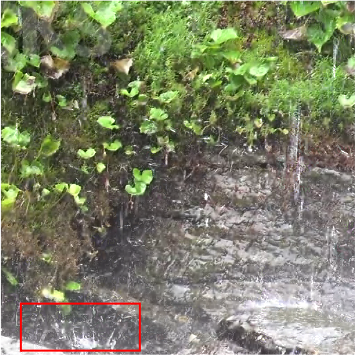}
\\
Rainy Image
\end{tabular}
\end{adjustbox}
\hspace{-0.1cm}
\begin{adjustbox}{valign=t}
\begin{tabular}{cccccc}
\includegraphics[width=0.196\textwidth]{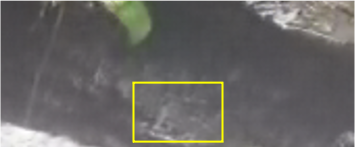} \hspace{-1mm} &
\includegraphics[width=0.196\textwidth]{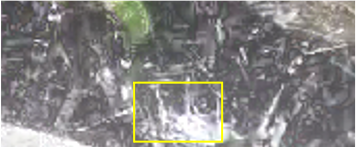} \hspace{-1mm} &
\includegraphics[width=0.196\textwidth]{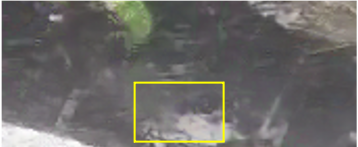} \hspace{-1mm} &
\includegraphics[width=0.196\textwidth]{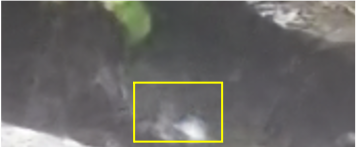} \hspace{-1mm} 
\\
Reference \hspace{-1mm} &
Rainy \hspace{-1mm} &
RCDNet~\cite{wang2020model} \hspace{-1mm} &
Restormer~\cite{zamir2022restormer} \hspace{-1mm} 
\\
\includegraphics[width=0.196\textwidth]{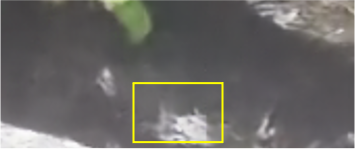} \hspace{-1mm} &
\includegraphics[width=0.196\textwidth]{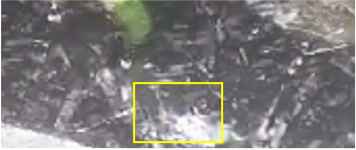} \hspace{-1mm} &
\includegraphics[width=0.196\textwidth]{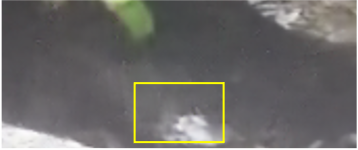} \hspace{-1mm} &
\includegraphics[width=0.196\textwidth]{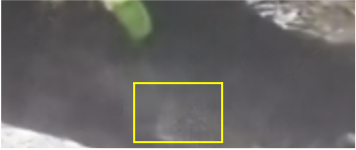} \hspace{-1mm} 
\\ 
IDT~\cite{xiao2022image} \hspace{-1mm} &
SPANet~\cite{wang2019spatial} \hspace{-1mm} &
Uformer~\cite{wang2022uformer} \hspace{-1mm} &
\textbf{HIT-B} \hspace{-1mm}
\\
\end{tabular}
\end{adjustbox}
\end{tabular}
\vspace{-2mm}
\caption{Qualitative comparisons with SOTA methods on SPAD~\cite{wang2019spatial} for deraining. }
\label{pic:derain}
\vspace{-3.mm}
\end{figure*}

\begin{table}[!t]
\centering
\caption{Quantitative comparison on {SIDD}~\cite{ssid_2018} and {DND}~\cite{dnd_2017}  for image denoising.}
\vspace{-3mm}
\footnotesize
\setlength{\tabcolsep}{1.435mm}
{
\begin{tabular}{cccccccccccccccc}
\toprule[0.8pt]
\multicolumn{2}{c}{\multirow{2}{*}{}} 
&  \multicolumn{4}{c|}{\textbf{SIDD}~\cite{ssid_2018}} & \multicolumn{4}{c||}{\textbf{DND}~\cite{dnd_2017}} & \multicolumn{4}{c}{Average} \\ 
\multicolumn{2}{l|}{\textbf{Method}} 
& \multicolumn{2}{c}{PSNR~$\uparrow$}& \multicolumn{2}{c|}{SSIM~$\uparrow$} & \multicolumn{2}{c}{PSNR~$\uparrow$} & \multicolumn{2}{c||}{SSIM~$\uparrow$}  & \multicolumn{2}{c}{PSNR~$\uparrow$} & \multicolumn{2}{c}{SSIM~$\uparrow$}  \\ 
\midrule[0.8pt]
    \multicolumn{2}{l|}{SDAP~\cite{Pan_2023_ICCV}}   
  & \multicolumn{2}{c}{37.53}& \multicolumn{2}{c|}{0.936} & \multicolumn{2}{c}{38.56} & \multicolumn{2}{c||}{0.940} & \multicolumn{2}{c}{38.05} & \multicolumn{2}{c}{0.938}\\
\multicolumn{2}{l|}{RIDNet~\cite{Anwar_2019_ICCV}}  
  & \multicolumn{2}{c}{38.71}& \multicolumn{2}{c|}{0.914} & \multicolumn{2}{c}{39.26} & \multicolumn{2}{c||}{0.953}  & \multicolumn{2}{c}{38.99} & \multicolumn{2}{c}{0.934}   \\

\multicolumn{2}{l|}{IPT~\cite{chen2021pre}}           
  & \multicolumn{2}{c}{39.10}& \multicolumn{2}{c|}{0.954} & \multicolumn{2}{c}{39.62} & \multicolumn{2}{c||}{0.952} & \multicolumn{2}{c}{39.36} & \multicolumn{2}{c}{0.953} \\
    \multicolumn{2}{l|}{VDN~\cite{yue2019variational}}          
  & \multicolumn{2}{c}{39.28}& \multicolumn{2}{c|}{0.909} & \multicolumn{2}{c}{39.38} & \multicolumn{2}{c||}{0.952}& \multicolumn{2}{c}{39.33} & \multicolumn{2}{c}{0.931} \\
    \multicolumn{2}{l|}{MalleNet~\cite{MalleableConvolution_eccv22}}    
  & \multicolumn{2}{c}{39.56}& \multicolumn{2}{c|}{0.941} & \multicolumn{2}{c}{39.21} & \multicolumn{2}{c||}{{0.949}}  & \multicolumn{2}{c}{39.39} & \multicolumn{2}{c}{0.945} \\
  \multicolumn{2}{l|}{MSANet~\cite{gou2022multiscale}}  
  & \multicolumn{2}{c}{39.56}& \multicolumn{2}{c|}{0.912} & \multicolumn{2}{c}{39.65} & \multicolumn{2}{c||}{\underline{0.955}} & \multicolumn{2}{c}{39.61} & \multicolumn{2}{c}{0.934}  \\
  \multicolumn{2}{l|}{VIRNet~\cite{yue2024deep}}   
  & \multicolumn{2}{c}{39.64}& \multicolumn{2}{c|}{0.958} & \multicolumn{2}{c}{39.83} & \multicolumn{2}{c||}{ 0.954} & \multicolumn{2}{c}{39.74} & \multicolumn{2}{c}{0.956}\\
    \multicolumn{2}{l|}{MPRNet~\cite{zamir2021multi}}   
  & \multicolumn{2}{c}{39.71}& \multicolumn{2}{c|}{0.958} & \multicolumn{2}{c}{39.80} & \multicolumn{2}{c||}{0.954} & \multicolumn{2}{c}{39.76} & \multicolumn{2}{c}{0.956}\\

      \multicolumn{2}{l|}{MIRNet-v2~\cite{MIRNetv2}}  
  & \multicolumn{2}{c}{39.84}& \multicolumn{2}{c|}{\underline{0.959}} & \multicolumn{2}{c}{39.86} & \multicolumn{2}{c||}{\underline{0.955}} & \multicolumn{2}{c}{\underline{39.85}} & \multicolumn{2}{c}{\underline{0.957}}\\
      \multicolumn{2}{l|}{DIL
      ~\cite{DIL_2023_CVPR}} 
  & \multicolumn{2}{c}{\underline{39.92}}& \multicolumn{2}{c|}{0.939} & \multicolumn{2}{c}{39.03} & \multicolumn{2}{c||}{\underline{0.955}} & \multicolumn{2}{c}{39.48} & \multicolumn{2}{c}{0.947}\\



\midrule
\multicolumn{2}{l|}{{\text{HIT-T (Ours)}}}     
  & \multicolumn{2}{c}{39.62}& \multicolumn{2}{c|}{0.958} & \multicolumn{2}{c}{\underline{39.93}} & \multicolumn{2}{c||}{\textbf{0.956}} & \multicolumn{2}{c}{{39.78}} & \multicolumn{2}{c}{\underline{0.957}}\\
\multicolumn{2}{l|}{{\text{HIT-B (Ours)}}}     
  & \multicolumn{2}{c}{\underline{39.94}}& \multicolumn{2}{c|}{\textbf{0.960}} & \multicolumn{2}{c}{\textbf{40.00}} & \multicolumn{2}{c||}{\textbf{0.956}} & \multicolumn{2}{c}{\underline{39.97}} & \multicolumn{2}{c}{\textbf{0.958}}\\


\bottomrule[0.8pt]
\end{tabular}}
\label{table4denoise}
\vspace{-2.5mm}
\end{table}
\noindent\textbf{Architecture Variants.}
By setting different feature channels C and the number of the Transformer blocks in the 4-level encoder-decoder architecture, we build two variants of HIT: 
1) HIT-T sets the feature channels to 16 and the number of Transformer blocks to [2, 2, 2, 2]; 
2) HIT-B sets the feature channels to 32 and the number of Transformer blocks to [1, 2, 8, 8]. 
In all experiments, the split window size is 8, and Transformer blocks share the same attention heads as~\cite{wang2022uformer}.

\noindent\textbf{Implementation Details.}
We train the model using AdamW optimizer ~\cite{loshchilov2017decoupled} 
with the recommended parameter settings from \cite{wang2022uformer}. 
The initial learning rate is set as $2e^{- 4}$ and gradually decreased to $1e^{- 6}$ using the cosine decay strategy. 
For data augmentation, we randomly adopt horizontal and vertical flips to the training samples. 
We adopt the progressive learning strategy to our model, similar to~\cite{zamir2022restormer,eccv2022_Stripformer}. 
%
%
\textbf{The code is provided in the \textcolor{magenta}{supplementary materials} to ensure the reproducibility of our results.}

\begin{table*}[!t]
\begin{minipage}{.48\linewidth}
\caption{Quantitative comparison on {SPAD} \cite{wang2019spatial} for image deraining.}
\vspace{-3mm}
\centering
\footnotesize
\scalebox{1.0}
{
\begin{tabular}{cccccccc}
\toprule[0.8pt]
\multicolumn{2}{c}{\multirow{2}{*}{}} 
& \multicolumn{4}{c}{\textbf{SPAD}~\cite{wang2019spatial}}\\ 
\multicolumn{2}{l|}{\textbf{Method}} 
& \multicolumn{2}{c}{PSNR~$\uparrow$}& \multicolumn{2}{c}{SSIM~$\uparrow$}   \\ \midrule[0.8pt]


  \multicolumn{2}{l|}{RESCAN~\cite{li2018recurrent}}                       
  & \multicolumn{2}{c}{38.11}& \multicolumn{2}{c}{0.9797}  \\

    \multicolumn{2}{l|}{SPANet~\cite{wang2019spatial}}                       
  & \multicolumn{2}{c}{40.24}& \multicolumn{2}{c}{0.9811}  \\
    \multicolumn{2}{l|}{RCDNet~\cite{wang2020model}}                       
  & \multicolumn{2}{c}{43.36}& \multicolumn{2}{c}{0.9831}  \\
      \multicolumn{2}{l|}{SPAIR~\cite{purohit2021spatially}}             
  & \multicolumn{2}{c}{44.10}& \multicolumn{2}{c}{0.9872}  \\
    \multicolumn{2}{l|}{Fu \etal~\cite{10035447}}                       
  & \multicolumn{2}{c}{45.03}& \multicolumn{2}{c}{0.9907}  \\
  
  \multicolumn{2}{l|}{Uformer~\cite{wang2022uformer}}                       
  & \multicolumn{2}{c}{46.13}& \multicolumn{2}{c}{0.9913}    \\
  \multicolumn{2}{l|}{Restormer~\cite{zamir2022restormer}}                       
  & \multicolumn{2}{c}{46.25}& \multicolumn{2}{c}{0.9911}  \\
  \multicolumn{2}{l|}{SCD-Former~\cite{Guo_2023_ICCV}}   
  & \multicolumn{2}{c}{46.89}& \multicolumn{2}{c}{\textbf{0.9941}}    \\
\multicolumn{2}{l|}{IDT~\cite{xiao2022image}}         
  & \multicolumn{2}{c}{47.34}& \multicolumn{2}{c}{0.9929}    \\
        \multicolumn{2}{l|}{DRSformer~\cite{DRSformer}}       
  & \multicolumn{2}{c}{\underline{48.53}}& \multicolumn{2}{c}{0.9924}  \\

 \midrule
\multicolumn{2}{l|}{{HIT-T (Ours)}}           
  & \multicolumn{2}{c}{{47.16}}& \multicolumn{2}{c}{{0.9926}}  \\
  \multicolumn{2}{l|}{{HIT-B (Ours)}}           
  & \multicolumn{2}{c}{\underline{49.16}}& \multicolumn{2}{c}{\underline{0.9940}}  \\
\bottomrule[0.8pt]
\end{tabular}}
\label{table4derain}
\vspace{-3.5mm}
\end{minipage}~~~~~\begin{minipage}{.48\linewidth}
\caption{Quantitative comparison on SMID~\cite{chen2019seeing} for image enhancement. 
}
\vspace{-3.mm}
\centering
\footnotesize
\scalebox{1.0}
{
\begin{tabular}{cccccccccc}
\toprule[0.8pt]
\multicolumn{2}{l}{\textbf{}}   & \multicolumn{4}{c}{\textbf{SMID~\cite{chen2019seeing}}
}\\ 
\multicolumn{2}{l|}{\textbf{Method}}  & \multicolumn{2}{c}{{PSNR~$\uparrow$}}  & \multicolumn{2}{c}{{SSIM~$\uparrow$}}\\  
\midrule[0.8pt]

\multicolumn{2}{l|}{KinD \cite{zhang2019kindling}}  & \multicolumn{2}{c}{22.18} & \multicolumn{2}{c}{0.634}\\ 
\multicolumn{2}{l|}{EnlightenGAN \cite{jiang2021enlightengan}}  & \multicolumn{2}{c}{22.62} & \multicolumn{2}{c}{0.674}\\ 
\multicolumn{2}{l|}{RetineNet~\cite{Chen2018Retinex}}  & \multicolumn{2}{c}{22.83} & \multicolumn{2}{c}{0.684}\\ 

\multicolumn{2}{l|}{DeepUPE \cite{wang2019underexposed}}  & \multicolumn{2}{c}{23.91} & \multicolumn{2}{c}{0.690}\\ 
\multicolumn{2}{l|}{SID~\cite{chen2018learning}}  & \multicolumn{2}{c}{24.78} & \multicolumn{2}{c}{0.718}\\ 
\multicolumn{2}{l|}{RUAS \cite{liu2021retinex}}  & \multicolumn{2}{c}{25.88} & \multicolumn{2}{c}{0.744}\\ 
\multicolumn{2}{l|}{Restormer \cite{zamir2022restormer}}  & \multicolumn{2}{c}{26.97} & \multicolumn{2}{c}{0.758}\\
\multicolumn{2}{l|}{Uformer~\cite{wang2022uformer}}  & \multicolumn{2}{c}{27.20} & \multicolumn{2}{c}{0.792}\\
\multicolumn{2}{l|}{SNR-Net~\cite{xu2022snr}}  & \multicolumn{2}{c}{28.49} & \multicolumn{2}{c}{0.805}\\
\multicolumn{2}{l|}{Retinexformer~\cite{retinexformer}}  & \multicolumn{2}{c}{29.15} & \multicolumn{2}{c}{\underline{0.815}}\\
\midrule[0.8pt]
\multicolumn{2}{l|}{HIT-T (Ours)}  & \multicolumn{2}{c}{\underline{29.16}}  & \multicolumn{2}{c}{{0.813}}\\ 
\multicolumn{2}{l|}{HIT-B (Ours)}  & \multicolumn{2}{c}{\textbf{29.37}}  & \multicolumn{2}{c}{\underline{0.821}}\\ 

\bottomrule[0.8pt]
\end{tabular}}
\label{tab:enhancement}
\vspace{-3.5mm}
\end{minipage}
\end{table*}
\subsection{Image Denosing}
We compare HIT with ten state-of-the-art (SOTA) denoising methods: SDAP~\cite{Pan_2023_ICCV}, RIDNet~\cite{Anwar_2019_ICCV}, IPT~\cite{chen2021pre}, VDN~\cite{yue2019variational}, 
MalleNet~\cite{MalleableConvolution_eccv22}, MSANet~\cite{gou2022multiscale}, VIRNet~\cite{yue2024deep}, MPRNet~\cite{zamir2021multi}, MIRNet-v2~\cite{MIRNetv2} and DIL~\cite{DIL_2023_CVPR}. 
Table~\ref{table4denoise} shows the quantitative result on the {SIDD}~\cite{ssid_2018} and {DND}~\cite{dnd_2017} benchmarks. 
%
%
It is noted that HIT-B trained on SIDD dataset not only obtains better performance (39.94 dB) on the same dataset than the SOTA (\eg, DIL~\cite{DIL_2023_CVPR}), but also makes a clear gain (0.97 dB) on DND dataset, which demonstrates its better generalization capability. 
Figure~\ref{pic:denoise} shows that HIT-B effectively removes noise while keeping image details well. 

\subsection{Image Deraining}
We compare HIT with ten SOTA deraining methods:
RESCAN~\cite{li2018recurrent},  SPANet~\cite{wang2019spatial}, RCDNet~\cite{wang2020model}, SPAIR~\cite{purohit2021spatially}, 
Fu \etal~\cite{10035447}, Uformer~\cite{wang2022uformer}, Restormer~\cite{zamir2022restormer}, SCD-Former~\cite{Guo_2023_ICCV}, IDT~\cite{xiao2022image} and DRSformer~\cite{DRSformer}. 
Table~\ref{table4derain} shows the quantitative results of HIT-B 
on SPAD~\cite{wang2019spatial} benchmark. 
HIT-B achieves a performance boost of 4.13 dB over the recent approach~\cite{10035447} and 0.63 dB over the previous best method DRSformer~\cite{DRSformer}. 
%
%
Figure~\ref{pic:derain} shows that HIT-B restores a visually better image.

\subsection{Low-Light Image Enhancement}
\label{sec:enhance}
We compare HIT with ten SOTA methods for low-light image enhancement: KinD \cite{zhang2019kindling}, EnlightenGAN \cite{jiang2021enlightengan}, RetineNet~\cite{Chen2018Retinex}, DeepUPE \cite{wang2019underexposed}, SID~\cite{chen2018learning}, RUAS \cite{liu2021retinex}, Restormer \cite{zamir2022restormer}, Uformer~\cite{wang2022uformer}, SNR-Net~\cite{xu2022snr} and Retinexformer~\cite{retinexformer}.
We report quantitative results on the SMID~\cite{chen2019seeing} dataset in Table \ref{tab:enhancement}.
Our HIT-B obtains the best performance among all the compared methods in terms of PSNR and SSIM metrics.
HIT-B makes a substantial performance gain of 0.22 dB when compared to the previous best Retinex-based method Retinexformer~\cite{retinexformer}.

\begin{table*}[t]
\caption{Quantitative comparison on Dense-Haze \cite{ancuti2019dense} for real image dehazing.
}
\vspace{-3.mm}

\centering
\scalebox{0.66}
{
\begin{tabular}{ccccccccccccc}
\toprule[0.8pt]
 \multicolumn{1}{c|}{\textbf{Method}} & \multicolumn{1}{c}{DCP}& \multicolumn{1}{c}{SGID} & \multicolumn{1}{c}{AOD-Net}& \multicolumn{1}{c}{FFA-Net} & \multicolumn{1}{c}{Uformer}& \multicolumn{1}{c}{Restormer} & \multicolumn{1}{c}{AECR-Net}& \multicolumn{1}{c}{Fourmer} & \multicolumn{1}{c}{DeHamer}& \multicolumn{1}{c}{MB-TaylorFormer} & \multicolumn{1}{|c}{HIT-T}& \multicolumn{1}{c}{HIT-B}
 \\ 
\multicolumn{1}{c|}{} & \multicolumn{1}{c}{\cite{he2010single}}& \multicolumn{1}{c}{\cite{bai2022self}} & \multicolumn{1}{c}{\cite{li2017aod}}& \multicolumn{1}{c}{\cite{qin2020ffa}} & \multicolumn{1}{c}{\cite{wang2022uformer}}& \multicolumn{1}{c}{\cite{zamir2022restormer}} & \multicolumn{1}{c}{\cite{wu2021contrastive}}& \multicolumn{1}{c}{\cite{zhou2023fourmer}} & \multicolumn{1}{c}{\cite{Guo_2022_CVPR}}& \multicolumn{1}{c}{\cite{qiu2023mb}} & \multicolumn{1}{|c}{(Ours)}& \multicolumn{1}{c}{(Ours)} \\ 
\hline
\multicolumn{1}{c|}{PSNR~$\uparrow$} & \multicolumn{1}{c}{10.06}& \multicolumn{1}{c}{13.09} & \multicolumn{1}{c}{13.14}& \multicolumn{1}{c}{14.39} & \multicolumn{1}{c}{15.22}& \multicolumn{1}{c}{15.78} & \multicolumn{1}{c}{15.80}& \multicolumn{1}{c}{15.95} & \multicolumn{1}{c}{16.62}& \multicolumn{1}{c}{\underline{16.66}} & \multicolumn{1}{|c}{{15.93}}& \multicolumn{1}{c}{\textbf{17.06}}
\\ 
 \multicolumn{1}{c|}{SSIM~$\uparrow$} & \multicolumn{1}{c}{~~0.39}& \multicolumn{1}{c}{~~0.52} & \multicolumn{1}{c}{~~0.41}& \multicolumn{1}{c}{~~0.45} & \multicolumn{1}{c}{~~0.43}& \multicolumn{1}{c}{~~\underline{0.55}} & \multicolumn{1}{c}{~~0.47}& \multicolumn{1}{c}{~~0.49} & \multicolumn{1}{c}{~~\textbf{0.56}}& \multicolumn{1}{c}{~~\textbf{0.56}} & \multicolumn{1}{|c}{~~{0.50}}& \multicolumn{1}{c}{~~\textbf{0.56}}
 \\ 
\bottomrule[0.8pt]
\end{tabular}}
\label{table4dehaze_appdix}
\end{table*}

\begin{table}[t]\footnotesize
\caption{Quantitative comparison on {RealBlur}~\cite{eccv2020real} for image debluring. 
All methods are only trained on GoPro~\cite{cvpr2017_gopro}.
}
\vspace{-3.mm}
\centering
\footnotesize
\setlength{\tabcolsep}{.685mm}{
\begin{tabular}{cccccccccccccc}
\toprule[0.8pt]
\multicolumn{2}{c}{\multirow{2}{*}{}} & 
\multicolumn{4}{c|}{\textbf{RealBlur-R}~\cite{eccv2020real}} & \multicolumn{4}{c}{\textbf{RealBlur-J}~\cite{eccv2020real}}& \multicolumn{4}{||c}{Average}  
\\ 
\multicolumn{2}{l|}{\textbf{Method}} & 
\multicolumn{2}{c}{PSNR~$\uparrow$}& \multicolumn{2}{c|}{SSIM~$\uparrow$} & \multicolumn{2}{c}{PSNR~$\uparrow$} & \multicolumn{2}{c}{SSIM~$\uparrow$}& \multicolumn{2}{||c}{PSNR~$\uparrow$} & \multicolumn{2}{c}{SSIM~$\uparrow$} 
\\\midrule[0.8pt]


\multicolumn{2}{l|}{IR-SDE~
\cite{luo2023image}}& \multicolumn{2}{c}{32.56} & \multicolumn{2}{c|}{0.909} & \multicolumn{2}{c}{23.19}  & \multicolumn{2}{c}{0.691} & \multicolumn{2}{||c}{27.89} & \multicolumn{2}{c}{0.800} \\
\multicolumn{2}{l|}{NAFNet~\cite{chen2022simple}}
  &  \multicolumn{2}{c}{33.63} & \multicolumn{2}{c|}{0.944} & \multicolumn{2}{c}{26.33} & \multicolumn{2}{c}{\underline{0.856}}& \multicolumn{2}{||c}{29.98} & \multicolumn{2}{c}{0.900}  
  \\
\multicolumn{2}{l|}{FFTformer~\cite{kong2023efficient}}
  &  \multicolumn{2}{c}{33.66} & \multicolumn{2}{c|}{\underline{0.948}} & \multicolumn{2}{c}{25.71} & \multicolumn{2}{c}{0.851}& \multicolumn{2}{||c}{29.69} & \multicolumn{2}{c}{0.900}  
  \\
\multicolumn{2}{l|}{CODE~
 \cite{Zhao_2023_CVPR}}                     & \multicolumn{2}{c}{33.81}& \multicolumn{2}{c|}{0.939} & \multicolumn{2}{c}{26.25} & \multicolumn{2}{c|}{0.801}  & \multicolumn{2}{||c}{30.03} & \multicolumn{2}{c}{0.870}\\
\multicolumn{2}{l|}{GRL-B~\cite{li2023grl}}                
  & \multicolumn{2}{c}{33.97} & \multicolumn{2}{c|}{0.944} & \multicolumn{2}{c}{{26.40}} & \multicolumn{2}{c}{0.816}& \multicolumn{2}{||c}{30.19} & \multicolumn{2}{c}{0.880}  
  \\
 \midrule
  \multicolumn{2}{l|}{{}{HIT-T} (Ours)} 
  & \multicolumn{2}{c}{\underline{35.23}}& \multicolumn{2}{c|}{{0.946}} & \multicolumn{2}{c}{\underline{28.36}
  } & \multicolumn{2}{c}{{0.855}}   & \multicolumn{2}{||c}{\underline{31.81}} & \multicolumn{2}{c}{\underline{0.901}} \\
    \multicolumn{2}{l|}{{}{HIT-B} (Ours)} 
  & \multicolumn{2}{c}{\textbf{36.19}}& \multicolumn{2}{c|}{\textbf{0.956}} & \multicolumn{2}{c}{\textbf{28.69}} & \multicolumn{2}{c}{\textbf{0.870}}   & \multicolumn{2}{||c}{\textbf{32.44}} & \multicolumn{2}{c}{\textbf{0.913}} \\
\bottomrule[0.8pt]
\end{tabular}}
\label{tab:deblur_real}
\vspace{-3.5mm}
\end{table}

\subsection{Image Dehazing}
We compare HIT with ten SOTA dehazing methods, including DCP \cite{he2010single}, SGID \cite{bai2022self}, AOD-Net \cite{li2017aod}, FFA-Net \cite{qin2020ffa}, Uformer \cite{wang2022uformer}, AECR-Net \cite{wu2021contrastive}, Restormer\cite{zamir2022restormer}, Fourmer\cite{zhou2023fourmer}, DeHamer \cite{Guo_2022_CVPR} and MB-TaylorFormer \cite{qiu2023mb}. 
As shown in Table~\ref{table4dehaze_appdix}, HIT-B achieves the best scores in PSNR. 
Specifically, HIT-B obtains a performance gain of 0.4 dB over the previous best method MB-TaylorFormer~\cite{qiu2023mb}, and 1.11 dB over the recent method Fourmer~\cite{zhou2023fourmer}. 
%
%

\subsection{Image Deblurring}
%
%
We provide the comparisons with five state-of-the-art techniques, including IR-SDE~\cite{luo2023image}, NAFNet~\cite{chen2022simple}, FFTformer~\cite{kong2023efficient}, CODE~\cite{Zhao_2023_CVPR} and GRL-B~\cite{li2023grl} on the RealBlur~\cite{eccv2020real} benchmark in Table~\ref{tab:deblur_real}. 
Specifically, HIT-B surpasses the recent method FFTformer~\cite{kong2023efficient} and GRL-B~\cite{li2023grl} by 2.75 dB and 2.25 dB, respectively. 
It should be noted that our model is trained only on the GoPro dataset, while HIT-B achieves competitive performance on the RealBlur benchmark, indicating its better generalization capability.

\subsection{Model Efficiency}
We present a detailed analysis of HIT against five state-of-the-art architectures that are designed for image deraining, including 
MPRNet \cite{zamir2021multi}, SwinIR~\cite{iccv2021_swinIR}, Uformer \cite{wang2022uformer}, Restormer \cite{zamir2022restormer}, IDT~\cite{xiao2022image}, and DRSformer~\cite{DRSformer}. 
%
%
As shown in Table~\ref{tab:deblur_compare}, HIT-B performs the best PSNR metric among all the considered methods. 
%
%
%
%
It is clear that our model achieves a good trade-off between restoration ability and computational cost. 
On the one hand, Uformer~\cite{wang2022uformer} and Restormer~\cite{zamir2022restormer} take the common solution, manipulate convolutional operations, to enrich local features and make a clear performance improvement over SwinIR~\cite{iccv2021_swinIR}. 
Since the core self-attention mechanism in Transformer hinders these models from achieving the satisfactory ability to capture local patterns, such a strategy is insufficient for obtaining adequate local information as ours. 
HIT leverages much richer high-frequence information, aided by the proposed modules, and performs better than all the compared methods. 
\begin{table*}[t]
\caption{Model efficiency analysis on SPAD~\cite{wang2019spatial}.}
\vspace{-3.mm}
\centering
\scalebox{.75}
{
\begin{tabular}{ccccccccc}
\toprule[0.8pt]
\multicolumn{1}{c|}{{Method}} & \multicolumn{1}{c|}{MPRNet~\cite{zamir2021multi}}& \multicolumn{1}{c|}{SwinIR~\cite{iccv2021_swinIR}} & \multicolumn{1}{c|}{Uformer-S~\cite{wang2022uformer}} & \multicolumn{1}{c|}{Restormer~\cite{zamir2022restormer}}& \multicolumn{1}{c|}{IDT~\cite{xiao2022image}} & \multicolumn{1}{c|}{DRSformer~\cite{DRSformer}}& \multicolumn{1}{c}{HIT-T}& \multicolumn{1}{c}{HIT-B}
\\ 
\midrule[0.8pt]
\multicolumn{1}{c|}{FLOPs/G} & \multicolumn{1}{c}{175.8} & \multicolumn{1}{c}{238.0} & \multicolumn{1}{c}{\underline{43.9}} & \multicolumn{1}{c}{174.7}& \multicolumn{1}{c}{{61.9}}& \multicolumn{1}{c|}{242.9}& \multicolumn{1}{c}{\textbf{15.8}}& \multicolumn{1}{c}{93.8}%
\\
\multicolumn{1}{c|}{Run-times/s}& \multicolumn{1}{c}{\textbf{0.03}} & \multicolumn{1}{c}{1.83}  & \multicolumn{1}{c}{0.12}&\multicolumn{1}{c}{0.14}&\multicolumn{1}{c}{0.28}&\multicolumn{1}{c|}{{0.08}}&\multicolumn{1}{c}{\underline{0.06}}&\multicolumn{1}{c}{{0.09}}
\\
\multicolumn{1}{c|}{PSNR/dB}& \multicolumn{1}{c}{43.64} & \multicolumn{1}{c}{44.97}  & \multicolumn{1}{c}{46.13}&\multicolumn{1}{c}{46.25}&\multicolumn{1}{c}{47.34}&\multicolumn{1}{c|}{\underline{48.53}}&\multicolumn{1}{c}{{47.16}}&\multicolumn{1}{c}{\textbf{49.16}}%
\\
\bottomrule[0.8pt]
\end{tabular}}
\label{tab:deblur_compare}
\end{table*}

\section{Ablation Studies}
To better understand the effect of each component, we provide ablation studies and train all possible baselines on the SPAD~\cite{wang2019spatial} for fair comparisons. 
GMACs are calculated by an input with the size of 256 $\times$ 256. 
The conclusions based on image deraining hold on other tasks. 
Due to the limited space, more detailed analysis and discussions are included in the supplemental material. 
%

\noindent\textbf{Window-wise Injection Module~(WIM).} 
%
To demonstrate the effectiveness of the proposed WIM, we conduct in-depth ablation experiments by comparing it with other components proposed for the same fusion purposes in other computer vision tasks. 
Specifically, we consider (b) the Feature Coupling Unit~(FCU) from \cite{iccv2021conformer}, (c) TransUNet from \cite{chen2021transunet}, (d) TCM from \cite{Liu_2023_CVPR_TCM} and (e) Mobile$\rightarrow$Former from \cite{cvpr2022mobile} as replacements for WIM in our experiments.

\begin{figure*}[t]
\begin{minipage}{.48\linewidth}
\tiny
\centering
\begin{tabular}{ccc}
\hspace{-0.26cm}
\begin{adjustbox}{valign=t}
\begin{tabular}{ccc}
\includegraphics[width=0.32\textwidth]{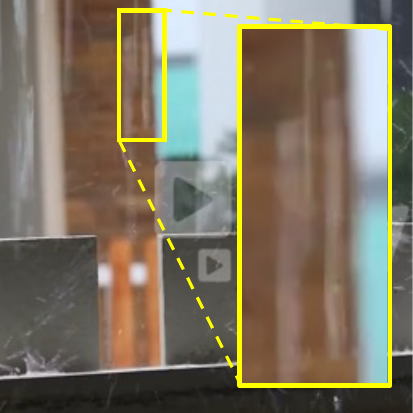} \hspace{-1mm} &
\includegraphics[width=0.32\textwidth]{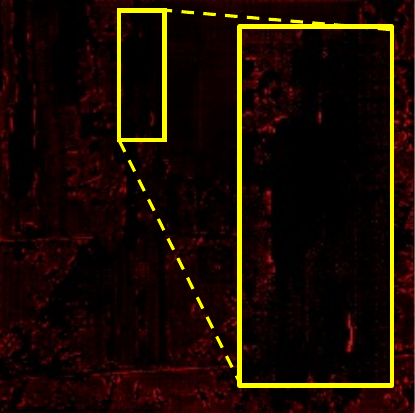} \hspace{-1mm} &
\includegraphics[width=0.32\textwidth]{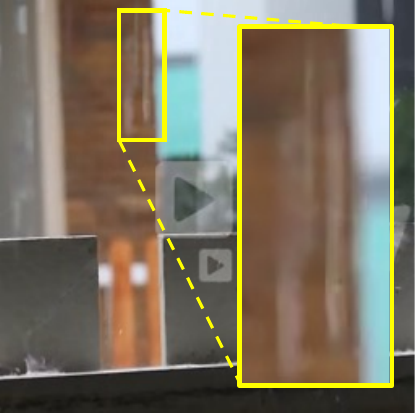} \hspace{-1mm} 
\\
(a)Rainy\hspace{-1mm} &
(b)Res. w/o WIM\hspace{-1mm} &
(c)Target w/o WIM\hspace{-1mm} 
\\
\includegraphics[width=0.32\textwidth]{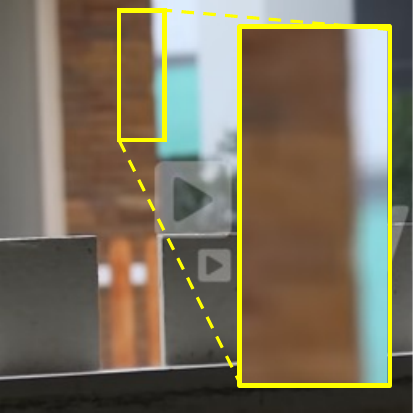} \hspace{-1mm} &
\includegraphics[width=0.32\textwidth]{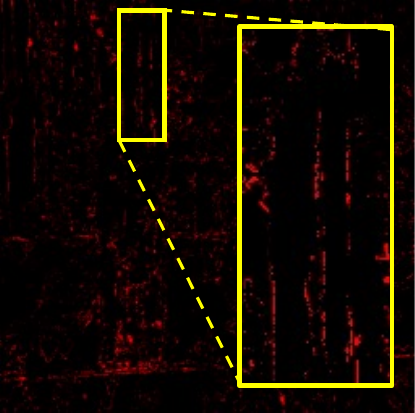} \hspace{-1mm} &
\includegraphics[width=0.32\textwidth]{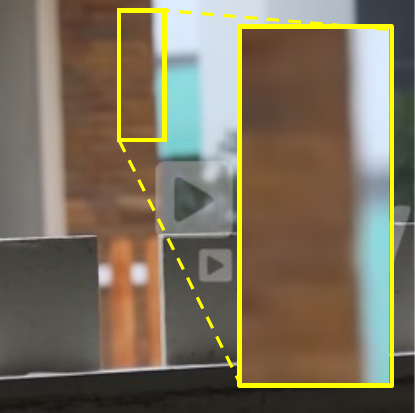} \hspace{-1mm} 
\\ 
(d)Reference\hspace{-1mm} &
(e)Res. w/ WIM\hspace{-1mm} &
(f)Target w/ WIM\hspace{-1mm} 
\\
\end{tabular}
\end{adjustbox}
\end{tabular}
\vspace{-3.mm}
\caption{Effect of the Window-wise Injection Module~(WIM). Compared with (b), HIT w/ WIM (e) can restore much more details in the residual result. 
Compared with (c), HIT w/ WIM generates a visually clearer result (f) and is closer to the reference one~(d).}
\label{pic:resdual_wem}
\vspace{-1.5mm}
\end{minipage}~~~~~~~~\begin{minipage}{.48\linewidth}
\vspace{-3.5mm}
\tiny
\centering
\begin{tabular}{ccc}
\hspace{-0.26cm}
\begin{adjustbox}{valign=t}
\begin{tabular}{ccc}
\includegraphics[width=0.32\textwidth]{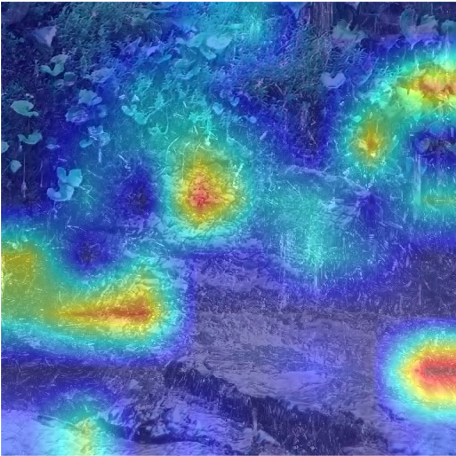} \hspace{-1mm} &
\includegraphics[width=0.32\textwidth]{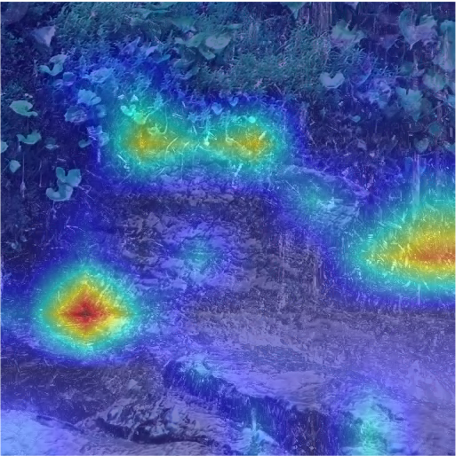} \hspace{-1mm} &
\includegraphics[width=0.32\textwidth]{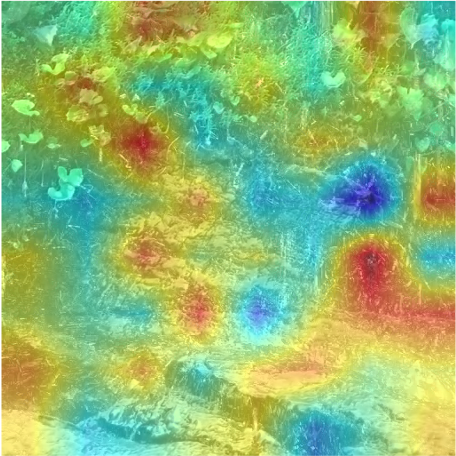} \hspace{-1mm} 
\\

\\
\includegraphics[width=0.32\textwidth]{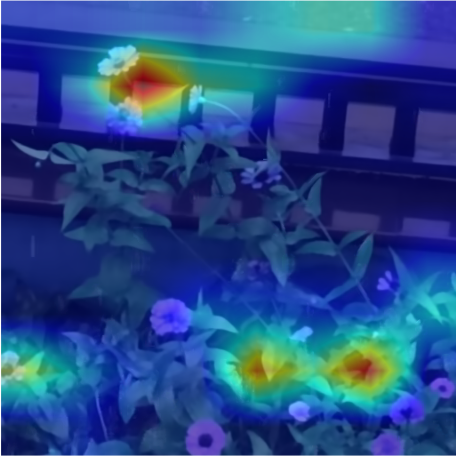} \hspace{-1mm} &
\includegraphics[width=0.32\textwidth]{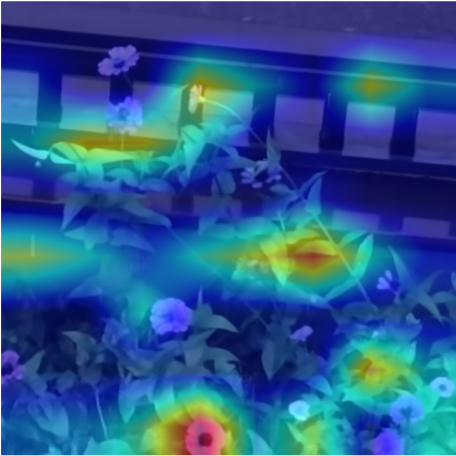} \hspace{-1mm} &
\includegraphics[width=0.32\textwidth]{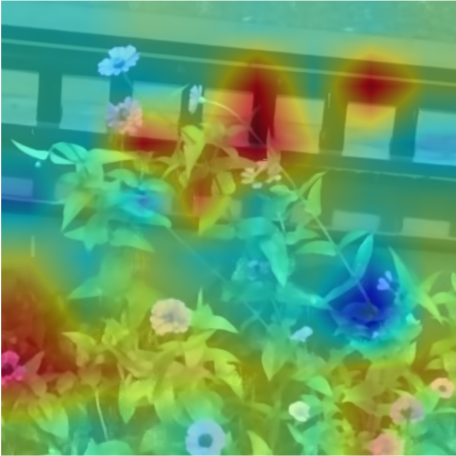} \hspace{-1mm} 
\\ 
(a) $\bm{F}_l$ \hspace{-1mm} &
(b) $\bm{F}_{l+1}$ \hspace{-1mm} &
(c) $\bm{F}_{sl}$ \hspace{-1mm} 
\\
\end{tabular}
\end{adjustbox}
\end{tabular}
\vspace{-3mm}
\caption{Feature analysis of Bidirectional Interaction Module (BIM). 
Instead of directly delivering the feature~(a), BIM aggregates features at different scales~((a) and (b)) and enhances the representation~(c) as a result. 
} 
\label{pic:cfm_vis}
\vspace{-1.5mm}
\end{minipage}
\end{figure*}

As shown in Table~\ref{tab:deblur-ablation}, using the FCU (Table~\ref{tab:deblur-ablation}b) leads to a performance drop~(a substantial 2.66 dB).
And when WIM is changed to TransUNet (Table~\ref{tab:deblur-ablation}c), we observe a more decline~(3.63 dB). 
This can be attributed to that the core component (\ie, Batch Normalization) in these designs is less suitable for image restoration~\cite{cvpr2017_gopro}. 
In addition, if Mobile$\rightarrow$Former (Table~\ref{tab:deblur-ablation}d) is adopted, which also serves as an unsuitable component, a drop of 1.64 dB in performance occurs. 
This decrease may be triggered by a significant reduction in computational capacity, which is required for deployment on mobile devices. 
Besides, though the model equipped with TCM (Table~\ref{tab:deblur-ablation}e) achieves the best scores among the considered varieties, there remains a clear performance (0.58 dB) gap with our WIM. 
%
%
Compared to these methods, it is noteworthy that WIM is the only one that splits input features into windows, aligning them explicitly with the downstream transformer branch. 
This design choice allows for high-frequency details preserved within each window that are exploited by the subsequent attention mechanism. 
In conclusion, even though these related works introduce various modules to combine the CNN and Transformer block, few of them consider the indispensable role of high-frequency information for image restoration. 
Moreover, we present visual comparisons in Figure~\ref{pic:resdual_wem} to demonstrate the effectiveness of the WIM.
It improves the model’s ability to handle degradation patterns, resulting in a cleaner restored image and more details in the residual image.

\begin{table}[t]
\caption{Ablation study of HIT-T on SPAD~\cite{wang2019spatial}. 
}
\vspace{-3.mm}
\centering
\footnotesize
\setlength{\tabcolsep}{1.4855mm}
{
\begin{tabular}{cccccccccc}
\toprule[0.8pt]
\multicolumn{2}{c|}{\textbf{Ablation}} &\multicolumn{2}{c|}{\textbf{Variant}} & \multicolumn{2}{c|}{\textbf{Param}} & \multicolumn{2}{c|}{\textbf{GMACs}}&  \multicolumn{2}{c}{\textbf{PSNR}} \\ \midrule[0.8pt]

\multicolumn{2}{c|}{None} &   \multicolumn{2}{l|}{(a) HIT-T (Ours)}    & \multicolumn{2}{c|}{17.29M}& \multicolumn{2}{c|}{15.76G} &  \multicolumn{2}{c}{42.98} \\
\midrule
\multicolumn{2}{c|}{\multirow{4}{*}{WIM}} &   \multicolumn{2}{l|}{(b) FCU~\cite{iccv2021conformer}}    & \multicolumn{2}{c|}{17.29M
}& \multicolumn{2}{c|}{
15.76G
} &  \multicolumn{2}{c}{40.32} \\
\multicolumn{2}{c|}{\multirow{2}{*}{}} &   \multicolumn{2}{l|}{(c) TransUNet~\cite{chen2021transunet}}    & \multicolumn{2}{c|}{17.30M
}& \multicolumn{2}{c|}{16.40G
} &  \multicolumn{2}{c}{39.35} \\
\multicolumn{2}{c|}{} &   \multicolumn{2}{l|}{(d) Mobile$\rightarrow$Former~\cite{cvpr2022mobile}}    &\multicolumn{2}{c|}{17.30M
}& \multicolumn{2}{c|}{
20.70G
} &  \multicolumn{2}{c}{41.34} \\
\multicolumn{2}{c|}{\multirow{2}{*}{}} &   \multicolumn{2}{l|}{(e) TCM~\cite{Liu_2023_CVPR_TCM}}    & \multicolumn{2}{c|}{17.29M
}& \multicolumn{2}{c|}{
15.80G
} &  \multicolumn{2}{c}{42.40} \\

\midrule
\multicolumn{2}{c|}{\multirow{3}{*}{BIM}} &   \multicolumn{2}{l|}{(f) Concat~\cite{ronneberger2015u}}    &\multicolumn{2}{c|}{16.81M}& \multicolumn{2}{c|}{
13.61G
} &  \multicolumn{2}{c}{41.79} \\
\multicolumn{2}{c|}{} &   \multicolumn{2}{l|}{(g) AFF~\cite{Chen_2021_ICCV}}    &\multicolumn{2}{c|}{17.12M}& \multicolumn{2}{c|}{
15.18G
} &  \multicolumn{2}{c}{41.84} \\
\multicolumn{2}{c|}{} &   \multicolumn{2}{l|}{(h) MDTA~\cite{zamir2022restormer}}    &\multicolumn{2}{c|}{17.29M}& \multicolumn{2}{c|}{
15.90G
} &  \multicolumn{2}{c}{42.69} \\
\midrule
\multicolumn{2}{c|}{\multirow{2}{*}{SEU}} &   \multicolumn{2}{l|}{(i) w/o SEU}    & \multicolumn{2}{c|}{17.22M}& \multicolumn{2}{c|}{
15.51G
} &  \multicolumn{2}{c}{41.04} \\
\multicolumn{2}{c|}{} &   \multicolumn{2}{l|}{(j) DPE~\cite{li2023uniformer}}    &\multicolumn{2}{c|}{17.22M}& \multicolumn{2}{c|}{
15.54G
} &  \multicolumn{2}{c}{40.48} \\
\bottomrule
\end{tabular}}
\label{tab:deblur-ablation}
\end{table}
\noindent\textbf{Bidirectional Interaction Module (BIM).} 
%
To validate the effectiveness of BIM, we compare it with three baselines: (f) a simple concatenation operation; (g) Asymmetric feature fusion (AFF) in \cite{cvpr2021cho}; (h) canceling cross query paradigm in BIM, which degrades to MDTA \cite{zamir2022restormer} style. 
%
Specifically, BIM (Table~\ref{tab:deblur-ablation}a), AFF (Table~\ref{tab:deblur-ablation}g), and MDTA version of BIM (Table~\ref{tab:deblur-ablation}h) all consider taking advantage of features at different scales to improve the presentations, and they enjoy clear benefit on performance over the simple concatenation (Table~\ref{tab:deblur-ablation}f). 
%
%
Since the features from different scales could emphasize distinct semantic information, directly merging them may lead to a semantic conflict issue, which can be aggravated as the number of features increases. 
Compared to handling all available features (\textgreater2) in AFF (Table~\ref{tab:deblur-ablation}g), coping with features from only two scales potentially introduces less semantic confusion (Table~\ref{tab:deblur-ablation}a and Table~\ref{tab:deblur-ablation}h). 
Furthermore, BIM introduces a bidirectional query paradigm to achieve the alignment of the features semantically and enhance the feature representations. 
In brief, all these dedicated designs made in BIM result in a simple yet effective solution to aggregate features at different scales. 
In addition, our visualization of the feature map processed by BIM (specifically, the final convolution layer in BIM), as shown in Figure~\ref{pic:cfm_vis}, clearly illustrates the efficacy of the proposed BIM. 
%
%
In our case, the final convolution layer in BIM is regarded as the target layer, and the real size of the visualized feature map is $H \times W$. 
Features at different scales are explored to facilitate the discriminate capability of the deeper attention layer. 

\noindent\textbf{Spatial Enhancement Unit (SEU).}
To evaluate the effectiveness of SEU, we conduct ablation studies by (i) canceling SEU and (j) replacing it with DPE~\cite{li2023uniformer}. 
We observe a clear performance degradation when SEU is not used in BIM (Table~\ref{tab:deblur-ablation}i), and replacing SEU with DPE~\cite{li2023uniformer} also leads to a significant reduction (Table~\ref{tab:deblur-ablation}j). 
These results demonstrate that SEU plays a necessary role in retaining key spatial information by executing in parallel to the attention calculation, whereas DPE precedes this step. 
DPE adopts a convolution layer like SEU and results in spatially enhanced features, however, these semantically different features with enhanced spatial information may challenge the model to learn satisfactory representations. 
Meantime, the spatial information could be lost during subsequent channel-wise attention computation within BIM, which leads to a significant performance drop (Table~\ref{tab:deblur-ablation}j).

\begin{table*}[t]\footnotesize
\caption{Quantitative comparison on {SPAD} \cite{wang2019spatial} for real image deraining.}
\vspace{-3mm}
\centering
\footnotesize
\scalebox{0.7250}
{
\begin{tabular}{cccccccccccccccc}
\toprule[0.8pt]
\multicolumn{2}{l|}{\textbf{Method}} & \multicolumn{2}{c}{SPAIR~\cite{purohit2021spatially}}   &  \multicolumn{2}{c}{Fu \etal~\cite{10035447}} & \multicolumn{2}{c}{Uformer-S~\cite{wang2022uformer}}&\multicolumn{2}{c}{Restormer~\cite{zamir2022restormer}}& \multicolumn{2}{c|}{Uformer-B~\cite{wang2022uformer}}&\multicolumn{2}{c}{IDT~\cite{xiao2022image}} &\multicolumn{2}{c}{IDT~\cite{xiao2022image}+WIM+BIM}\\ 
\multicolumn{2}{l|}{Reference} &\multicolumn{2}{c}{ICCV'21}   & \multicolumn{2}{c}{TPAMI'23} & \multicolumn{2}{c}{CVPR'22}&\multicolumn{2}{c}{CVPR'22}& \multicolumn{2}{c|}{CVPR'22}  &\multicolumn{2}{c}{TPAMI'22} &\multicolumn{2}{c}{TPAMI'22}  \\ 
\midrule[0.8pt]
\multicolumn{2}{l|}{PSNR~$\uparrow$} &\multicolumn{2}{c}{44.10}    & \multicolumn{2}{c}{45.03} & \multicolumn{2}{c}{46.13}& \multicolumn{2}{c}{46.25}& \multicolumn{2}{c|}{\underline{47.84}}  &\multicolumn{2}{c}{47.34} &\multicolumn{2}{c}{\textbf{47.91}}  \\
\multicolumn{2}{l|}{SSIM~$\uparrow$} &\multicolumn{2}{c}{0.9872}   & \multicolumn{2}{c}{0.9907} & \multicolumn{2}{c}{0.9913}& \multicolumn{2}{c}{0.9917}& \multicolumn{2}{c|}{0.9925}  &\multicolumn{2}{c}{\underline{0.9929}} &\multicolumn{2}{c}{\textbf{0.9935}}  \\

\bottomrule[0.8pt]
\end{tabular}}
\label{table4derain_appdix}
\vspace{-3mm}
\end{table*}

\noindent\textbf{Extension to new Baseline.}
To further demonstrate the effectiveness of our approach in injecting high-frequency information within the Transformer for image restoration, we conduct extension experiments. 
Specifically, we implement WIM and BIM on the recent work IDT~\cite{xiao2022image} for image deraining. 
As shown in Table~\ref{table4derain_appdix}, we observe a clear improvement in both PSNR and SSIM metrics~(\eg, 0.57 dB on PSNR metric). 
With the assistance of our modules, the IDT model surpasses the previous SOTA method {Uformer-B}~\cite{wang2022uformer}.

\section{Conclusion}
In this paper, we present a new Transformer-based model (HIT) with injected high-frequency information. 
We develop the window-wise injection module and bidirectional interaction module to help the Transformer benefit from the crucial local cues when analyzing the degraded patterns. 
We show that HIT can handle a variety of image restoration tasks including denoising, deraining, deblurring, demoiréing, deraindrop, dehazing, desnowing, deshadowing, and low-light enhancement, and performs competitively in terms of computational cost and accuracy.

%
%
\bibliographystyle{splncs04}
\bibliography{HIT}

\begin{thebibliography}{100}
\providecommand{\url}[1]{\texttt{#1}}
\providecommand{\urlprefix}{URL }
\providecommand{\doi}[1]{https://doi.org/#1}

\bibitem{ssid_2018}
Abdelhamed, A., Lin, S., Brown, M.S.: A high-quality denoising dataset for smartphone cameras. In: CVPR (2018)

\bibitem{ancuti2019dense}
Ancuti, C.O., Ancuti, C., Sbert, M., Timofte, R.: Dense-haze: A benchmark for image dehazing with dense-haze and haze-free images. In: ICIP (2019)

\bibitem{Anwar_2019_ICCV}
Anwar, S., Barnes, N.: Real image denoising with feature attention. In: ICCV (2019)

\bibitem{bai2022self}
Bai, H., Pan, J., Xiang, X., Tang, J.: Self-guided image dehazing using progressive feature fusion. TIP  (2022)

\bibitem{retinexformer}
Cai, Y., Bian, H., Lin, J., Wang, H., Timofte, R., Zhang, Y.: Retinexformer: One-stage retinex-based transformer for low-light image enhancement. In: ICCV (2023)

\bibitem{charbonnier1994two}
Charbonnier, P., Blanc-Feraud, L., Aubert, G., Barlaud, M.: Two deterministic half-quadratic regularization algorithms for computed imaging. In: ICIP (1994)

\bibitem{chen2019seeing}
Chen, C., Chen, Q., Do, M.N., Koltun, V.: Seeing motion in the dark. In: ICCV (2019)

\bibitem{chen2018learning}
Chen, C., Chen, Q., Xu, J., Koltun, V.: Learning to see in the dark. In: CVPR (2018)

\bibitem{chen2021pre}
Chen, H., Wang, Y., Guo, T., Xu, C., Deng, Y., Liu, Z., Ma, S., Xu, C., Xu, C., Gao, W.: Pre-trained image processing transformer. In: CVPR (2021)

\bibitem{chen2021transunet}
Chen, J., Lu, Y., Yu, Q., Luo, X., Adeli, E., Wang, Y., Lu, L., Yuille, A.L., Zhou, Y.: Transunet: Transformers make strong encoders for medical image segmentation. arXiv preprint arXiv:2102.04306  (2021)

\bibitem{chen2022simple}
Chen, L., Chu, X., Zhang, X., Sun, J.: Simple baselines for image restoration. In: ECCV (2022)

\bibitem{Chen_2021_ICCV}
Chen, W.T., Fang, H.Y., Hsieh, C.L., Tsai, C.C., Chen, I.H., Ding, J.J., Kuo, S.Y.: All snow removed: Single image desnowing algorithm using hierarchical dual-tree complex wavelet representation and contradict channel loss. In: ICCV (2021)

\bibitem{DRSformer}
Chen, X., Li, H., Li, M., Pan, J.: Learning a sparse transformer network for effective image deraining. In: CVPR (2023)

\bibitem{cvpr2022mobile}
Chen, Y., Dai, X., Chen, D., Liu, M., Dong, X., Yuan, L., Liu, Z.: Mobile-former: Bridging mobilenet and transformer. In: CVPR (2022)

\bibitem{cvpr2021cho}
Cho, S.J., Ji, S.W., Hong, J.P., Jung, S.W., Ko, S.J.: Rethinking coarse-to-fine approach in single image deblurring. In: ICCV (2021)

\bibitem{cho2009fast}
Cho, S., Lee, S.: Fast motion deblurring. ACM TOG  (2009)

\bibitem{iclr2021_vit}
Dosovitskiy, A., Beyer, L., Kolesnikov, A., Weissenborn, D., Zhai, X., Unterthiner, T., Dehghani, M., Minderer, M., Heigold, G., Gelly, S., Uszkoreit, J., Houlsby, N.: An image is worth 16x16 words: Transformers for image recognition at scale. In: ICLR (2021)

\bibitem{d2021convit}
d’Ascoli, S., Touvron, H., Leavitt, M.L., Morcos, A.S., Biroli, G., Sagun, L.: Convit: Improving vision transformers with soft convolutional inductive biases. In: ICML (2021)

\bibitem{fan2020neural}
Fan, Y., Yu, J., Mei, Y., Zhang, Y., Fu, Y., Liu, D., Huang, T.S.: Neural sparse representation for image restoration. In: NeurIPS (2020)

\bibitem{fergus2006removing}
Fergus, R., Singh, B., Hertzmann, A., Roweis, S.T., Freeman, W.T.: Removing camera shake from a single photograph. ACM TOG  (2006)

\bibitem{10035447}
Fu, X., Xiao, J., Zhu, Y., Liu, A., Wu, F., Zha, Z.J.: Continual image deraining with hypergraph convolutional networks. TPAMI  (2023)

\bibitem{pami22_sr_noise}
Gao, S., Zhuang, X.: Rank-one network: An effective framework for image restoration. TPAMI  (2022)

\bibitem{gou2022multiscale}
Gou, Y., Hu, P., Lv, J., Zhou, J.T., Peng, X.: Multi-scale adaptive network for single image denoising. In: NeurIPS (2022)

\bibitem{Guo_2022_CVPR}
Guo, C.L., Yan, Q., Anwar, S., Cong, R., Ren, W., Li, C.: Image dehazing transformer with transmission-aware 3d position embedding. In: CVPR (2022)

\bibitem{guo2023spatial}
Guo, S., Yong, H., Zhang, X., Ma, J., Zhang, L.: Spatial-frequency attention for image denoising. arXiv preprint arXiv:2302.13598  (2023)

\bibitem{Guo_2023_ICCV}
Guo, Y., Xiao, X., Chang, Y., Deng, S., Yan, L.: From sky to the ground: A large-scale benchmark and simple baseline towards real rain removal. In: ICCV (2023)

\bibitem{MopNet}
He, B., Wang, C., Shi, B., Duan, L.: Mop moiré patterns using mopnet. In: ICCV (2019)

\bibitem{he2010single}
He, K., Sun, J., Tang, X.: Single image haze removal using dark channel prior. TPAMI  (2010)

\bibitem{pami23_rain_huang}
Huang, H., Luo, M., He, R.: Memory uncertainty learning for real-world single image deraining. TPAMI  (2023)

\bibitem{jiang2021enlightengan}
Jiang, Y., Gong, X., Liu, D., Cheng, Y., Fang, C., Shen, X., Yang, J., Zhou, P., Wang, Z.: Enlightengan: Deep light enhancement without paired supervision. TIP  (2021)

\bibitem{MalleableConvolution_eccv22}
Jiang, Y., Wronski, B., Mildenhall, B., Barron, J.T., Wang, Z., Xue, T.: Fast and high quality image denoising via malleable convolution. In: ECCV (2022)

\bibitem{pami22_noise_blur_kerihuan}
Ke, R., Schönlieb, C.B.: Unsupervised image restoration using partially linear denoisers. TPAMI  (2022)

\bibitem{kong2023efficient}
Kong, L., Dong, J., Ge, J., Li, M., Pan, J.: Efficient frequency domain-based transformers for high-quality image deblurring. In: CVPR (2023)

\bibitem{li2017aod}
Li, B., Peng, X., Wang, Z., Xu, J., Feng, D.: Aod-net: All-in-one dehazing network. In: ICCV (2017)

\bibitem{pami22_survey_enhancement}
Li, C., Guo, C., Han, L., Jiang, J., Cheng, M.M., Gu, J., Loy, C.C.: Low-light image and video enhancement using deep learning: A survey. TPAMI  (2022)

\bibitem{li2023uniformer}
Li, K., Wang, Y., Zhang, J., Gao, P., Song, G., Liu, Y., Li, H., Qiao, Y.: Uniformer: Unifying convolution and self-attention for visual recognition. TPAMI  (2023)

\bibitem{li2018recurrent}
Li, X., Wu, J., Lin, Z., Liu, H., Zha, H.: Recurrent squeeze-and-excitation context aggregation net for single image deraining. In: ECCV (2018)

\bibitem{DIL_2023_CVPR}
Li, X., Li, B., Jin, X., Lan, C., Chen, Z.: Learning distortion invariant representation for image restoration from a causality perspective. In: CVPR (2023)

\bibitem{li2023grl}
Li, Y., Fan, Y., Xiang, X., Demandolx, D., Ranjan, R., Timofte, R., Gool, V.L.: Efficient and explicit modelling of image hierarchies for image restoration. In: CVPR (2023)

\bibitem{li2021localvit}
Li, Y., Zhang, K., Cao, J., Timofte, R., Van~Gool, L.: Localvit: Bringing locality to vision transformers. arXiv preprint arXiv:2104.05707  (2021)

\bibitem{iccv2021_swinIR}
Liang, J., Cao, J., Sun, G., Zhang, K., Van~Gool, L., Timofte, R.: Swinir: Image restoration using swin transformer. In: ICCV Workshops (2021)

\bibitem{Liu_2023_CVPR_TCM}
Liu, J., Sun, H., Katto, J.: Learned image compression with mixed transformer-cnn architectures. In: CVPR (2023)

\bibitem{pami21_blur_liujun}
Liu, J., Yan, M., Zeng, T.: Surface-aware blind image deblurring. TPAMI  (2021)

\bibitem{TAPE-Net_eccv22}
Liu, L., Xie, L., Zhang, X., Yuan, S., Chen, X., Zhou, W., Li, H., Tian, Q.: Tape: Task-agnostic prior embedding for image restoration. In: ECCV (2022)

\bibitem{liu2021retinex}
Liu, R., Ma, L., Zhang, J., Fan, X., Luo, Z.: Retinex-inspired unrolling with cooperative prior architecture search for low-light image enhancement. In: CVPR (2021)

\bibitem{liu2018desnownet}
Liu, Y.F., Jaw, D.W., Huang, S.C., Hwang, J.N.: Desnownet: Context-aware deep network for snow removal. TIP  (2018)

\bibitem{liu2021swin}
Liu, Z., Lin, Y., Cao, Y., Hu, H., Wei, Y., Zhang, Z., Lin, S., Guo, B.: Swin transformer: Hierarchical vision transformer using shifted windows. In: ICCV (2021)

\bibitem{loshchilov2017decoupled}
Loshchilov, I., Hutter, F.: Decoupled weight decay regularization. arXiv preprint arXiv:1711.05101  (2017)

\bibitem{lu2022needs}
Lu, Y., Lin, Y., Wu, H., Luo, Y., Zheng, X., Wang, L.: All one needs to know about priors for deep image restoration and enhancement: A survey. arXiv preprint arXiv:2206.02070  (2022)

\bibitem{Lugmayr_2022_CVPR}
Lugmayr, A., Danelljan, M., Timofte, R., Kim, K.w., Kim, Y., Lee, J.y., Li, Z., Pan, J., Shim, D., Song, K.U., Tang, J., Wang, C., Zhao, Z.: Ntire 2022 challenge on learning the super-resolution space. In: CVPR Workshops (2022)

\bibitem{luo2023image}
Luo, Z., Gustafsson, F.K., Zhao, Z., Sj{\"o}lund, J., Sch{\"o}n, T.B.: Image restoration with mean-reverting stochastic differential equations. In: ICML (2023)

\bibitem{cvpr2017_gopro}
Nah, S., Hyun~Kim, T., Mu~Lee, K.: Deep multi-scale convolutional neural network for dynamic scene deblurring. In: CVPR (2017)

\bibitem{Nah_2021_CVPR}
Nah, S., Son, S., Lee, S., Timofte, R., Lee, K.M.: Ntire 2021 challenge on image deblurring. In: CVPR Workshops (2021)

\bibitem{pan2018learning}
Pan, J., Liu, S., Sun, D., Zhang, J., Liu, Y., Ren, J., Li, Z., Tang, J., Lu, H., Tai, Y.W., et~al.: Learning dual convolutional neural networks for low-level vision. In: CVPR (2018)

\bibitem{pan2017deblurring}
Pan, J., Sun, D., Pfister, H., Yang, M.H.: Deblurring images via dark channel prior. TPAMI  (2017)

\bibitem{pami22_panXinggang}
Pan, X., Zhan, X., Dai, B., Lin, D., Loy, C.C., Luo, P.: Exploiting deep generative prior for versatile image restoration and manipulation. TPAMI  (2022)

\bibitem{Pan_2023_ICCV}
Pan, Y., Liu, X., Liao, X., Cao, Y., Ren, C.: Random sub-samples generation for self-supervised real image denoising. In: ICCV (2023)

\bibitem{park2022how}
Park, N., Kim, S.: How do vision transformers work? In: ICLR (2022)

\bibitem{iccv2021conformer}
Peng, Z., Huang, W., Gu, S., Xie, L., Wang, Y., Jiao, J., Ye, Q.: Conformer: Local features coupling global representations for visual recognition. In: ICCV (2021)

\bibitem{dnd_2017}
Plotz, T., Roth, S.: Benchmarking denoising algorithms with real photographs. In: CVPR (2017)

\bibitem{potlapalli2023promptir}
Potlapalli, V., Zamir, S.W., Khan, S., Khan, F.S.: Promptir: Prompting for all-in-one blind image restoration. arXiv preprint arXiv:2306.13090  (2023)

\bibitem{purohit2021spatially}
Purohit, K., Suin, M., Rajagopalan, A., Boddeti, V.N.: Spatially-adaptive image restoration using distortion-guided networks. In: ICCV (2021)

\bibitem{qian2018attentive}
Qian, R., Tan, R.T., Yang, W., Su, J., Liu, J.: Attentive generative adversarial network for raindrop removal from a single image. In: CVPR (2018)

\bibitem{qin2020ffa}
Qin, X., Wang, Z., Bai, Y., Xie, X., Jia, H.: Ffa-net: Feature fusion attention network for single image dehazing. In: AAAI (2020)

\bibitem{qiu2023mb}
Qiu, Y., Zhang, K., Wang, C., Luo, W., Li, H., Jin, Z.: Mb-taylorformer: Multi-branch efficient transformer expanded by taylor formula for image dehazing. In: ICCV (2023)

\bibitem{pami22_blur_renWenqi}
Ren, W., Zhang, J., Pan, J., Liu, S., Ren, J.S., Du, J., Cao, X., Yang, M.H.: Deblurring dynamic scenes via spatially varying recurrent neural networks. TPAMI  (2022)

\bibitem{eccv2020real}
Rim, J., Lee, H., Won, J., Cho, S.: Real-world blur dataset for learning and benchmarking deblurring algorithms. In: ECCV (2020)

\bibitem{ronneberger2015u}
Ronneberger, O., Fischer, P., Brox, T.: U-net: Convolutional networks for biomedical image segmentation. In: MICCAI (2015)

\bibitem{tip23_songxibin}
Song, X., Zhou, D., Li, W., Dai, Y., Shen, Z., Zhang, L., Li, H.: Tusr-net: Triple unfolding single image dehazing with self-regularization and dual feature to pixel attention. TIP  (2023)

\bibitem{sun2015learning}
Sun, J., Cao, W., Xu, Z., Ponce, J.: Learning a convolutional neural network for non-uniform motion blur removal. In: CVPR (2015)

\bibitem{MSNet}
Sun, Y., Yu, Y., Wang, W.: Moiré photo restoration using multiresolution convolutional neural networks. TIP  (2018)

\bibitem{IG_17ICML}
Sundararajan, M., Taly, A., Yan, Q.: Axiomatic attribution for deep networks. In: ICML (2017)

\bibitem{eccv2022_Stripformer}
Tsai, F.J., Peng, Y.T., Lin, Y.Y., Tsai, C.C., Lin, C.W.: Stripformer: Strip transformer for fast image deblurring. In: ECCV (2022)

\bibitem{vaswani2017attention}
Vaswani, A., Shazeer, N., Parmar, N., Uszkoreit, J., Jones, L., Gomez, A.N., Kaiser, {\L}., Polosukhin, I.: Attention is all you need. In: NeurIPS (2017)

\bibitem{wang2020model}
Wang, H., Xie, Q., Zhao, Q., Meng, D.: A model-driven deep neural network for single image rain removal. In: CVPR (2020)

\bibitem{wang2018stacked}
Wang, J., Li, X., Yang, J.: Stacked conditional generative adversarial networks for jointly learning shadow detection and shadow removal. In: CVPR (2018)

\bibitem{wang2019underexposed}
Wang, R., Zhang, Q., Fu, C.W., Shen, X., Zheng, W.S., Jia, J.: Underexposed photo enhancement using deep illumination estimation. In: CVPR (2019)

\bibitem{wang2019spatial}
Wang, T., Yang, X., Xu, K., Chen, S., Zhang, Q., Lau, R.W.: Spatial attentive single-image deraining with a high quality real rain dataset. In: CVPR (2019)

\bibitem{pami23_wang_noise}
Wang, W., Wen, F., Yan, Z., Liu, P.: Optimal transport for unsupervised denoising learning. TPAMI  (2023)

\bibitem{wang2022uformer}
Wang, Z., Cun, X., Bao, J., Zhou, W., Liu, J., Li, H.: Uformer: A general u-shaped transformer for image restoration. In: CVPR (2022)

\bibitem{Chen2018Retinex}
Wei, C., Wang, W., Yang, W., Liu, J.: Deep retinex decomposition for low-light enhancement. In: BMVC (2018)

\bibitem{wu2021contrastive}
Wu, H., Qu, Y., Lin, S., Zhou, J., Qiao, R., Zhang, Z., Xie, Y., Ma, L.: Contrastive learning for compact single image dehazing. In: CVPR (2021)

\bibitem{xiao2022image}
Xiao, J., Fu, X., Liu, A., Wu, F., Zha, Z.J.: Image de-raining transformer. TPAMI  (2022)

\bibitem{xu2022snr}
Xu, X., Wang, R., Fu, C.W., Jia, J.: Snr-aware low-light image enhancement. In: CVPR (2022)

\bibitem{rdnet_tmm22}
Yue, H., Cheng, Y., Mao, Y., Cao, C., Yang, J.: Recaptured screen image demoiréing in raw domain. TMM  (2022)

\bibitem{yue2019variational}
Yue, Z., Yong, H., Zhao, Q., Meng, D., Zhang, L.: Variational denoising network: Toward blind noise modeling and removal. In: NeurIPS (2019)

\bibitem{yue2024deep}
Yue, Z., Yong, H., Zhao, Q., Zhang, L., Meng, D., Wong, K.Y.K.: Deep variational network toward blind image restoration. TPAMI  (2024)

\bibitem{zamir2022restormer}
Zamir, S.W., Arora, A., Khan, S., Hayat, M., Khan, F.S., Yang, M.H.: Restormer: Efficient transformer for high-resolution image restoration. In: CVPR (2022)

\bibitem{zamir2020cycleisp}
Zamir, S.W., Arora, A., Khan, S., Hayat, M., Khan, F.S., Yang, M.H., Shao, L.: Cycleisp: Real image restoration via improved data synthesis. In: CVPR (2020)

\bibitem{zamir2021multi}
Zamir, S.W., Arora, A., Khan, S., Hayat, M., Khan, F.S., Yang, M.H., Shao, L.: Multi-stage progressive image restoration. In: CVPR (2021)

\bibitem{MIRNetv2}
Zamir, S.W., Arora, A., Khan, S., Hayat, M., Khan, F.S., Yang, M.H., Shao, L.: Learning enriched features for fast image restoration and enhancement. TPAMI  (2023)

\bibitem{zeiler2014visualizing}
Zeiler, M.D., Fergus, R.: Visualizing and understanding convolutional networks. In: ECCV (2014)

\bibitem{zhang2021pami}
Zhang, K., Li, Y., Zuo, W., Zhang, L., Van~Gool, L., Timofte, R.: Plug-and-play image restoration with deep denoiser prior. TPAMI  (2021)

\bibitem{zhang2020deblurring}
Zhang, K., Luo, W., Zhong, Y., Ma, L., Stenger, B., Liu, W., Li, H.: Deblurring by realistic blurring. In: CVPR (2020)

\bibitem{zhang2022deep}
Zhang, K., Ren, W., Luo, W., Lai, W.S., Stenger, B., Yang, M.H., Li, H.: Deep image deblurring: A survey. IJCV  (2022)

\bibitem{zhang2019kindling}
Zhang, Y., Zhang, J., Guo, X.: Kindling the darkness: A practical low-light image enhancer. In: ACMMM (2019)

\bibitem{zhang2023realtime}
Zhang, Y., Lin, M., Li, X., Liu, H., Wang, G., Chao, F., Shuai, R., Wen, Y., Chen, X., Ji, R.: Real-time image demoireing on mobile devices. In: ICLR (2023)

\bibitem{Zhao_2023_CVPR}
Zhao, H., Gou, Y., Li, B., Peng, D., Lv, J., Peng, X.: Comprehensive and delicate: An efficient transformer for image restoration. In: CVPR (2023)

\bibitem{tpami22_zheng_demoire}
Zheng, B., Yuan, S., Yan, C., Tian, X., Zhang, J., Sun, Y., Liu, L., Leonardis, A., Slabaugh, G.: Learning frequency domain priors for image demoireing. TPAMI  (2022)

\bibitem{zheng2022cross}
Zheng, C., Zhang, Y., Gu, J., Zhang, Y., Kong, L., Yuan, X.: Cross aggregation transformer for image restoration. In: NeurIPS (2022)

\bibitem{zhou2023fourmer}
Zhou, M., Huang, J., Guo, C.L., Li, C.: Fourmer: an efficient global modeling paradigm for image restoration. In: ICML (2023)

\bibitem{zou2021sdwnet}
Zou, W., Jiang, M., Zhang, Y., Chen, L., Lu, Z., Wu, Y.: Sdwnet: A straight dilated network with wavelet transformation for image deblurring. In: ICCV (2021)

\end{thebibliography}
\end{document}